\documentclass[sigconf]{acmart}

\usepackage{booktabs} % For formal tables
\usepackage{graphicx}
\usepackage{multirow}
\usepackage{subfigure}
\usepackage{bm}
\usepackage{balance} 

\AtBeginDocument{%
  \providecommand\BibTeX{{%
    \normalfont B\kern-0.5em{\scshape i\kern-0.25em b}\kern-0.8em\TeX}}}

\copyrightyear{2020}
\acmYear{2020}
\setcopyright{acmlicensed}
\acmConference[MM '20]{Proceedings of the 28th ACM International Conference on Multimedia}{October 12--16, 2020}{Seattle, WA, USA}
\acmBooktitle{Proceedings of the 28th ACM International Conference on Multimedia (MM '20), October 12--16, 2020, Seattle, WA, USA}
\acmPrice{15.00}
\acmDOI{10.1145/3394171.3413967}
\acmISBN{978-1-4503-7988-5/20/10}

\begin{document}
\fancyhead{}

%%
%% The "title" command has an optional parameter,
%% allowing the author to define a "short title" to be used in page headers.
\title{Regularized Two-Branch Proposal Networks for Weakly-Supervised Moment Retrieval in Videos}

%%
%% The "author" command and its associated commands are used to define
%% the authors and their affiliations.
%% Of note is the shared affiliation of the first two authors, and the
%% "authornote" and "authornotemark" commands
%% used to denote shared contribution to the research.
% \author{Zhu Zhang}
% \authornote{Both authors contributed equally to this research.}
% \email{zhangzhu@zju.edu.cn}
% \orcid{1234-5678-9012}
% \author{Zhijie Lin}
% \authornotemark[1]
% \email{linzhijie@zju.edu.cn}
% \affiliation{%
%   \institution{Zhejiang University}
%   % \streetaddress{P.O. Box 1212}
%   % \city{Hangzhou}
%   % \country{China}
%   % \postcode{43017-6221}
% }

\author{Zhu Zhang}
\authornote{Both authors contributed equally to this research.}
\affiliation{%
  \institution{Zhejiang University}
  % \streetaddress{1 Th{\o}rv{\"a}ld Circle}
  % \city{Hangzhou}
  % \country{China}
}
\email{zhangzhu@zju.edu.cn}

\author{Zhijie Lin}
\authornotemark[1]
\affiliation{%
  \institution{Zhejiang University}
  % \streetaddress{P.O. Box 1212}
  % \city{Hangzhou}
  % \country{China}
  % \postcode{43017-6221}
}
\email{linzhijie@zju.edu.cn}

\author{Zhou Zhao}
\authornote{Zhou Zhao is the corresponding author.}
\affiliation{%
  \institution{Zhejiang University}
  % \streetaddress{1 Th{\o}rv{\"a}ld Circle}
  % \city{Hangzhou}
  % \country{China}
}
\email{zhaozhou@zju.edu.cn}

\author{Jieming Zhu}
\affiliation{%
  \institution{Huawei Noah's Ark Lab}
  % \city{Hangzhou}
  % \country{China}
}
\email{jamie.zhu@huawei.com}

\author{Xiuqiang He}
\affiliation{%
  \institution{Huawei Noah's Ark Lab}
  % \city{Hangzhou}
  % \country{China}
}
\email{hexiuqiang1@huawei.com}
% \author{Aparna Patel}
% \affiliation{%
%  \institution{Rajiv Gandhi University}
%  \streetaddress{Rono-Hills}
%  \city{Doimukh}
%  \state{Arunachal Pradesh}
%  \country{India}}

% \author{Huifen Chan}
% \affiliation{%
%   \institution{Tsinghua University}
%   \streetaddress{30 Shuangqing Rd}
%   \city{Haidian Qu}
%   \state{Beijing Shi}
%   \country{China}}

% \author{Charles Palmer}
% \affiliation{%
%   \institution{Palmer Research Laboratories}
%   \streetaddress{8600 Datapoint Drive}
%   \city{San Antonio}
%   \state{Texas}
%   \postcode{78229}}
% \email{cpalmer@prl.com}

%%
%% By default, the full list of authors will be used in the page
%% headers. Often, this list is too long, and will overlap
%% other information printed in the page headers. This command allows
%% the author to define a more concise list
%% of authors' names for this purpose.
\renewcommand{\shortauthors}{Trovato and Tobin, et al.}

%%
%% The abstract is a short summary of the work to be presented in the
%% article.
\begin{abstract}
Video moment retrieval aims to localize the target moment in an video according to the given sentence. The weak-supervised setting only provides the video-level sentence annotations during training. Most existing weak-supervised methods apply a MIL-based framework to develop inter-sample confrontment, but ignore the intra-sample confrontment between moments with semantically similar contents. Thus, these methods fail to distinguish the target moment from plausible negative moments. In this paper, we propose a novel Regularized Two-Branch Proposal Network to simultaneously consider the inter-sample and intra-sample confrontments. Concretely, we first devise a language-aware filter to generate an enhanced video stream and a suppressed video stream. We then design the sharable two-branch proposal module to generate positive proposals from the enhanced stream and plausible negative proposals from the suppressed one for sufficient confrontment. Further, we apply the proposal regularization to stabilize the training process and improve model performance. The extensive experiments show the effectiveness of our method. Our code is released at here\footnote{https://github.com/ikuinen/regularized\_two-branch\_proposal\_network}.
\end{abstract}

\ccsdesc[500]{Information systems~Video search}
\ccsdesc[500]{Computing methodologies~Activity recognition and understanding}

%%
%% Keywords. The author(s) should pick words that accurately describe
%% the work being presented. Separate the keywords with commas.
\keywords{Weakly-Supervised Moment Retrieval; Two-Branch; Regularization}

%% A "teaser" image appears between the author and affiliation
%% information and the body of the document, and typically spans the
%% page.
% \begin{teaserfigure}
%   \includegraphics[width=\textwidth]{sampleteaser}
%   \caption{Seattle Mariners at Spring Training, 2010.}
%   \Description{Enjoying the baseball game from the third-base
%   seats. Ichiro Suzuki preparing to bat.}
%   \label{fig:teaser}
% \end{teaserfigure}

%%
%% This command processes the author and affiliation and title
%% information and builds the first part of the formatted document.
\maketitle

\begin{figure}[t]
\centering
\includegraphics[width=1.0\columnwidth]{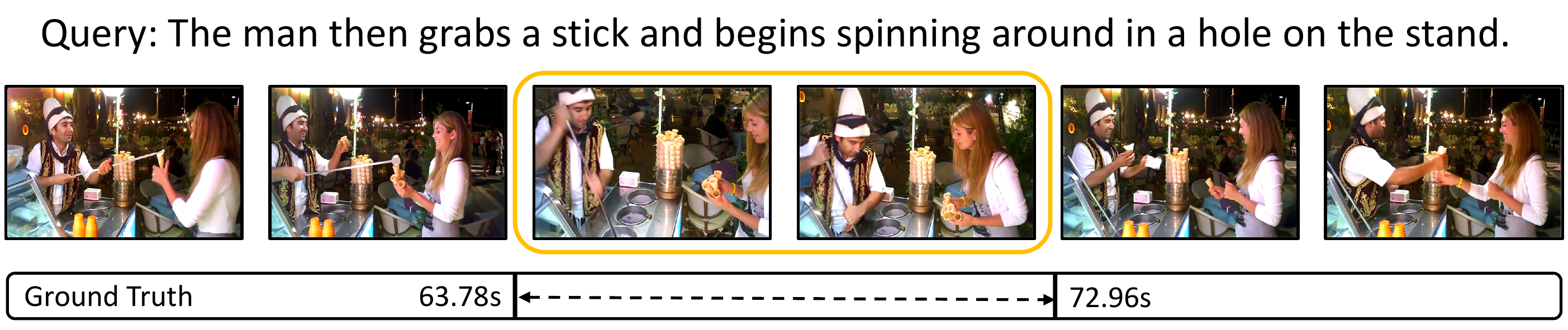} 
\caption{An example of video moment retrieval.}
\label{fig:example1}
\end{figure}

\section{Introduction}
Given a natural language description and an untrimmed video, video moment retrieval~\cite{gao2017tall,hendricks2017localizing} aims to automatically locate the temporal boundaries of the target moment semantically matching to the given sentence.
As shown in Figure 1, the sentence describes multiple complicated events and corresponds to a temporal moment with complex object interactions. 
Recently, a large amount of methods~\cite{gao2017tall,hendricks2017localizing,chen2018temporally,zhang2019cross,wang2019language} have been proposed to this challenging task and achieved satisfactory performance. 
However, most existing approaches are trained in the fully-supervised setting with the temporal alignment annotation of each sentence. Such manual annotations are very time-consuming and expensive, especially for ambiguous descriptions. But there is a mass of coarse descriptions for videos without temporal annotations on the Internet, such as the captions for videos on YouTube. Hence, in this paper, we develop a weakly-supervised method for video moment retrieval, which only needs the video-level sentence annotations rather than temporal boundary annotations for each sentence during training.

Most existing weakly-supervised moment retrieval works~\cite{mithun2019weakly,gao2019wslln,chen2020look} apply a Multiple Instance Learning (MIL)~\cite{karpathy2015deep} based methods. They regard matched video-sentence pairs as positive samples and unmatched video-sentence pairs as negative samples. Next, they learn the latent visual-textual alignment by inter-sample confrontment and utilize intermediate results to localize the target moment.
Concretely, Mithun et al.~\cite{mithun2019weakly} apply text-guided attention weights across frames to determine the reliant moment. 
And Gao and Chen et al.~\cite{gao2019wslln,chen2020look} measure the semantic consistency between texts and videos and then directly apply segment scores as localization clues.
However, these methods mainly focus on the inter-sample confrontment to judge whether the video matches with the given textual descriptions, but ignore the intra-sample confrontment to decide which moment matches the given language best.
Specifically, as shown in Figure 1, given a matched video-sentence pair, the video generally contains consecutive contents and these are a large amount of plausible negative moments, which have a bit of relevance to the language. It is intractable to distinguish the target moment from these plausible negative moments, especially when the plausible ones have large overlaps with the ground truth. Thus, we need to develop sufficient intra-sample confrontment between moments with similar contents in a video.

Based on above observations, we propose a novel Regularized Two-Branch Proposal Network~(RTBPN) to further explore the fine-grained intra-sample confrontment by discovering the plausible negative moment proposals.
Concretely, we first devise a language-aware filter to generate an enhanced video stream and a suppressed stream from the original video stream. In the enhanced stream, we highlight the critical frames according to the language information and weaken unnecessary ones. On the contrary, the crucial frames are suppressed in the suppressed stream. 
Next, we employ a two-branch proposal module to produce moment proposals from each stream, where the enhanced branch generates positive moment proposals and the suppressed branch produces plausible negative moment proposals. By the sufficient confrontment between two branches, we can accurately localize the most relevant moment from plausible ones.
But the suppressed branch may produce simple negative proposals rather than plausible ones, leading to ineffective confrontment. To avoid it, we share all parameters between two branches to make them possess the same ability to produce high-quality proposals. Moreover, parameter sharing can reduce network parameters and accelerate model convergence. By the two-branch framework, we can simultaneously develop sufficient inter-sample and intra-sample confrontment to boost the performance of weakly-supervised video moment retrieval.

Next, we consider the concrete design of the language-aware filter and two-branch proposal module. For the language-aware filter, we first project the language features into fixed cluster centers by a trainable generalized Vector of Locally Aggregated Descriptors (VLAD)~\cite{arandjelovic2016netvlad}, where each center can be regarded as a language scene, and then calculate the attention scores between scene and frame features as the language-to-frame relevance. Such a scene-based method introduces an intermediately semantic space for texts and videos, beneficial to boost the generalization ability. Next, to avoid producing a trivial score distribution, e.g. all frames are assigned to 1 or 0, we apply a max-min normalization on the distribution.
Based on the normalized distribution, we employ a two-branch gate to produce the enhanced and suppressed streams. 

As for the two-branch proposal module, two branches have a completely consistent structure and share all parameters. We first develop a conventional cross-modal interaction~\cite{chen2018temporally,zhang2019cross} between language and frame sequences. Next, we apply a 2D moment map~\cite{zhang2019learning} to capture relationships between adjacent moments. After it, we need to generate high-quality moment proposals from each branch. Most existing weakly-supervised approaches~\cite{mithun2019weakly,gao2019wslln,chen2020look} take all frames or moments as proposals to perform the inter-sample confrontment, which introduces a large amount of ineffective proposals into the training process.
Different from them, we devise a center-based proposal method to filter out unnecessary proposals and only retain high-quality ones. Specifically, we first determine the moment with the highest score as the center and then select those moments having high overlaps with the center one. This technique can effectively select a series of correlative moments to make the confrontment between two branches more sufficient.

Network regularization is widely-used in weakly-supervised tasks~\cite{liu2019completeness,chen2019weakly}, 
which injects extra limitations~(i.e. prior knowledge) into the network to stabilize the training process and improve the model performance. Here we design a proposal regularization strategy for our model, consisting of a global term and a gap term.
On the one hand, considering most of moments are semantically irrelevant to the language descriptions, we apply a global regularization term to make the average moment score relatively low, which implicitly encourages the scores of irrelevant moments close to 0.
On the other hand, we further expect to select the most accurate moment from positive moment proposals, thus we apply another gap regularization term to enlarge the score gaps between those positive moments for better identifying the target one.

Our main contributions can be summarized as follows:
\begin{itemize}
\item We design a novel Regularized Two-Branch Proposal Network for weakly-supervised video moment retrieval, which simultaneously considers the inter-sample and intra-sample confrontments by the sharable two-branch framework.
\item We devise the language-aware filter to generate the enhanced video stream and the suppressed one, and develop the sharable two-branch proposal module to produce the positive moment proposals and plausible negative ones for sufficient intra-sample confrontment.
\item We apply the proposal regularization strategy to stabilize the training process and improve the model performance.
\item The extensive experiments on three large-scale datasets show the effectiveness of our proposed RTBPN method.
\end{itemize}

\begin{figure*}[t]
    \centering
    \includegraphics[width=2.0\columnwidth]{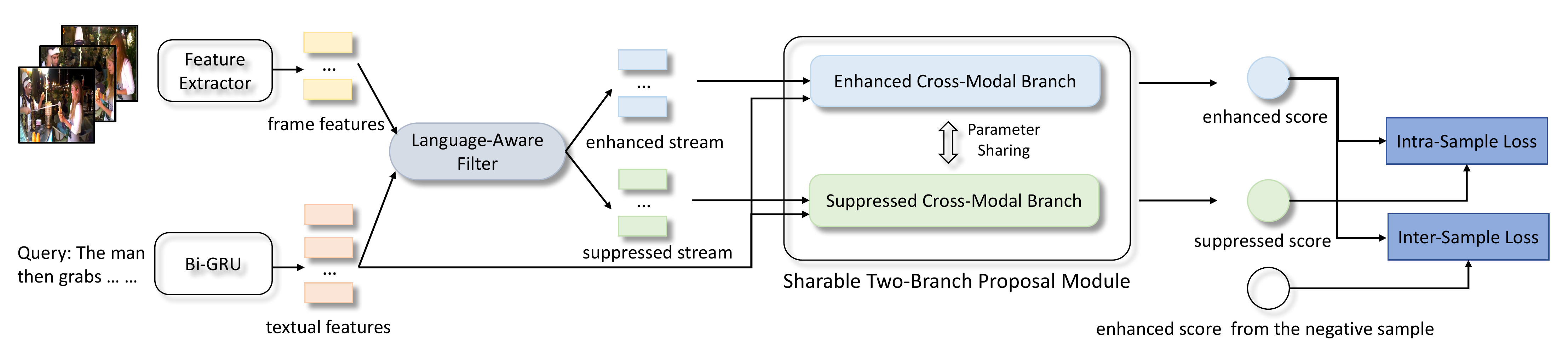} 
    \caption{The Overall Architecture of the Regularized Two-Branch Proposal Network. 
    % We first devise the language-aware filter to generate the enhanced video stream and the suppressed stream, and then develop the sharable two-branch proposal module to produce the positive moment proposals and plausible negative ones for sufficient intra-sample confrontment.
    }
    \label{fig:framework}
\end{figure*}

\section{Related Work}
% In this section, we briefly review some related works on temporal action localization and video moment retrieval.

\subsection{Temporal Action Localization}
Temporal action localization aims to detect the temporal boundaries and the categories of action instances in untrimmed videos.
The supervised methods~\cite{shou2016temporal,zhao2017temporal,shou2017cdc,chao2018rethinking,zeng2019graph} mainly adopt the two-stage framework, which first produces a series of temporal action proposals, then predicts the action class and regresses their boundaries. Concretely, Shou et al.~\cite{shou2016temporal} design three segment-based 3D ConvNet to accurately localize action instances and Zhao et al.~\cite{zhao2017temporal} apply a structured temporal pyramid to explore the context structure of actions. Recently, Chao et al.~\cite{chao2018rethinking} transfer the classical Faster-RCNN framework~\cite{ren2015faster} for action localization and Zeng et al.~\cite{zeng2019graph} exploit proposal-proposal relations using graph convolutional networks.

Under the weakly-supervised setting only with video-level action labels, Wang et al.~\cite{wang2017untrimmednets} design the classification and selection module to reason about the temporal duration of action instances. Nguyen et al.~\cite{nguyen2018weakly} utilize temporal class activations and class-agnostic attentions to localize the action segments. Further, Shou et al.~\cite{shou2018autoloc} propose a novel Outer-Inner-Contrastive loss to discover the segment-level supervision for action boundary prediction. To keep the completeness of actions, Liu et al.~\cite{liu2019completeness} employ a multi-branch framework where branches are enforced to discover distinctive parts of actions. And Yu et al.~\cite{yu2019temporal} explore the temporal action structure and model each action as a multi-phase process.

\subsection{Video Moment Retrieval}
Video moment retrieval aims to localize the target moment according to the given query in an untrimmed video. Most existing methods employ a top-down framework, which first generates a set of moment proposals and then selects the most relevant one. 
Early approaches~\cite{gao2017tall,hendricks2017localizing,hendricks2018localizing,liu2018attentive,liu2018cross} explicitly extract the moment proposals by the sliding windows with various lengths and individually calculate the correlation of each proposal with the query in a multi-modal space. To incorporate long-term video context, researchers~\cite{chen2018temporally,zhang2019man,zhang2019cross,lin2020moment,xu2019multilevel,zhang2019learning,yuan2019semantic} implicitly produce moment proposals by defining multiple temporal anchors after holistic visual-textual interactions. 
Concretely, Chen et al.~\cite{chen2018temporally} build sufficient frame-by-word interaction and dynamically aggregate the matching clues.
Zhang et al.~\cite{zhang2019man} employ an iterative graph adjustment network to learn moment-wise relations in a structured graph. 
% Yuan et al.~\cite{yuan2019semantic} propose a semantic conditioned dynamic modulation to better correlate and compose the sentence-related video contents over time.
And Zhang et al.~\cite{zhang2019learning} design a 2D temporal map to capture the temporal relations between adjacent moments.
Different from the top-down formula, the bottom-up framework~\cite{chen2019localizing,chenrethinking} is designed to directly predict the probabilities of each frame as target boundaries.
Further, He and Wang et al.~\cite{he2019read,wang2019language} formulate this task as a problem of sequential decision making and apply the reinforcement learning method to progressively regulate the temporal boundaries. 
Besides temporal moment retrieval, recent works~\cite{chen2019weakly,zhang2020does,zhang2020object} also localize the spatio-temporal tubes from videos according to the give language descriptions.
And Zhang et al.~\cite{zhang2019localizing} try to localize the target moment by the image query instead of the natural language query.

Recently, researchers~\cite{duan2018weakly,mithun2019weakly,gao2019wslln,lin2019weakly,chen2020look} begin to explore the weakly-supervised moment retrieval only with the video-level sentence annotations. 
% Duan et al.~\cite{duan2018weakly} take this task as the dual problem of weakly-supervised dense video captioning and develop a cycle system to train two tasks. 
Mithun, Gao and Chen et al.~\cite{mithun2019weakly,gao2019wslln,chen2020look} apply a MIL-based framework to learn latent visual-textual alignment by inter-sample confrontment.
Mithun et al.~\cite{mithun2019weakly} determine the reliant moment based on the intermediately text-guided attention weights.
Gao et al.~\cite{gao2019wslln} devise an alignment module to measure the semantic consistency between texts and videos and apply a detection module to compare moment proposals.
And Chen et al.~\cite{chen2020look} apply a two-stage model to detect the accurate moment in a coarse-to-fine manner.
Besides MIL-based methods, Lin et al.~\cite{lin2019weakly} propose a semantic completion network to rank proposals by a language reconstruction reward, but ignore the inter-sample confrontments.
Unlike previous methods, we design a sharable two-branch framework to simultaneously consider the inter-sample and intra-sample confrontments for weakly-supervised video moment retrieval.

\begin{figure*}[t]
    \centering
    \includegraphics[width=2.0\columnwidth]{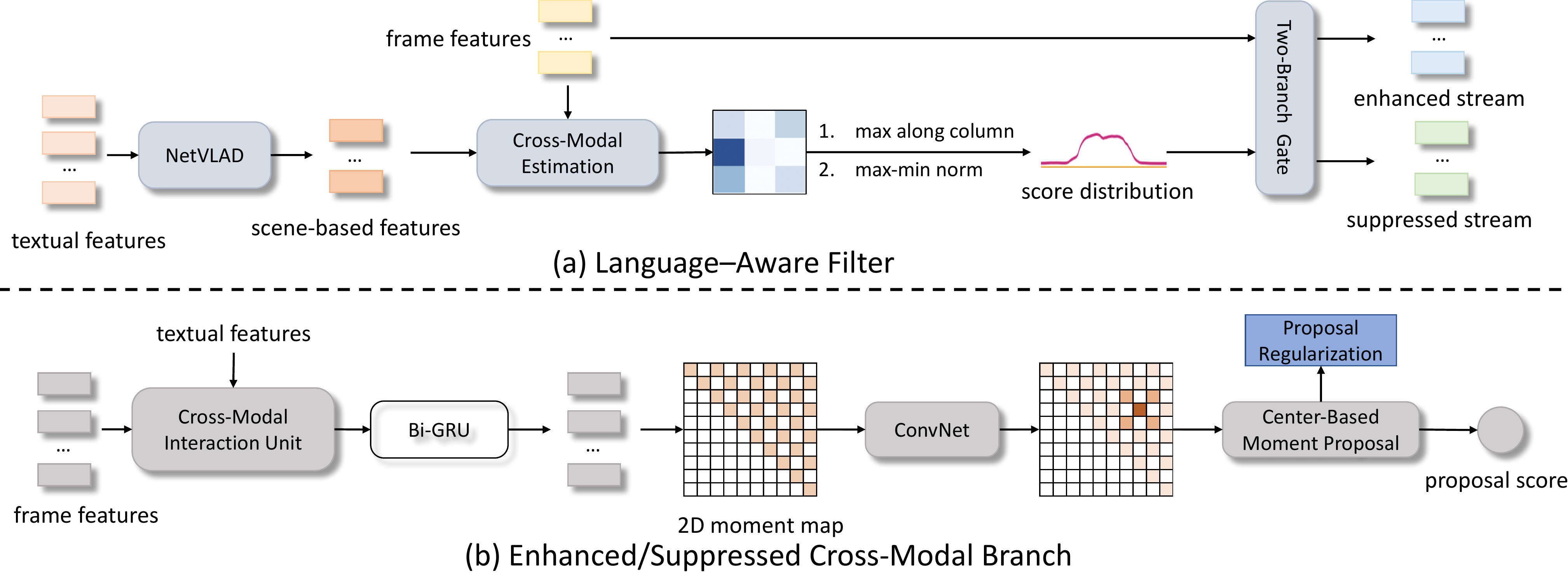} 
    \caption{ The Concrete Designs of the Language-Aware Filter and Sharable Two-Branch Proposal Module. 
    }
    \label{fig:framework2}
\end{figure*}

\section{The Proposed Method}
Given a video $V$ and a sentence $S$, video moment retrieval aims to retrieve the most relevant moment ${\hat l} = ({\hat s},{\hat e})$ within the video $V$, where ${\hat s}$ and ${\hat e}$ denote the indices of the start and end frames of the target moment. Due to the weakly-supervised setting, we can only utilize the coarse video-level annotations.
 % rather than temporal boundary annotations for each sentence during training.

\subsection{The Overall Architecture Design}
We first introduce the overall architecture of our Regularized Two-Branch Proposal Network (RTBPN). 
As shown in Figure~\ref{fig:framework}, we devise a language-aware filter to generate the enhanced video stream and the suppressed video stream, and next develop the sharable two-branch proposal module to produce the positive moment proposals and plausible negative ones. Finally, we develop the inter-sample and intra-sample losses with proposal regularization terms.

Concretely, we first extract the word features of the sentence by a pre-trained Glove word2vec embedding~\cite{pennington2014glove}. We then feed the word features into a Bi-GRU network~\cite{chung2014empirical} to learn word semantic representations ${\bf Q} = \{{\bf q}_i\}_{i=1}^{n_q}$ with contextual information, where $n_q$ is the word number and ${\bf q}_i$ is the semantic feature of the $i$-th word. As for videos, we first extract visual features using a pre-trained feature extractor (e.g. 3D-ConvNet~\cite{tran2015learning}) and then apply a temporal mean pooling to shorten the sequence length.  
We denote frame features as ${\bf V} = \{{\bf v}_i\}_{i=1}^{n_v}$, where $n_v$ is the feature number.

After feature extraction, we devise a language-aware filter to generate the enhanced and suppressed video streams, given by 
\begin{eqnarray}
& {\bf V}^{en}, {\bf V}^{sp} =  {\rm Filter} \ ({\bf V}, {\bf Q}),
\end{eqnarray}
where ${\bf V}^{en} = \{{\bf v}^{en}_i\}_{i=1}^{n_v}$ represents the enhanced video stream and ${\bf V}^{sp} = \{{\bf v}^{sp}_i\}_{i=1}^{n_v}$ is the suppressed video stream. In the enhance stream, we highlight the critical frame features relevant to the language and weaken unnecessary ones. On the contrary, the significative frames are suppressed in the suppressed stream. 

Next, we develop the sharable two-branch proposal module to produce the positive moment proposals and plausible negative ones. The module consists of a enhanced branch and a suppressed branch with the consistent structure and sharable parameters ${\Theta}$, given by
\begin{eqnarray}
\begin{aligned}
& {\rm P}^{en}, {\bf L}^{en}, {\bf C}^{en} =  {\rm EnhancedBranch}_{\Theta} \ ({\bf V}^{en}, {\bf Q}), \\
& {\rm P}^{sp}, {\bf L}^{sp}, {\bf C}^{sp} =  {\rm SuppressedBranch}_{\Theta} \ ({\bf V}^{sp}, {\bf Q}), \\
\end{aligned}
\end{eqnarray}
where we feed the enhanced video stream ${\bf V}^{en}$ and textual features ${\bf Q}$ into the enhanced branch and produce the positive moment proposals ${\rm P}^{en} = \{{p}^{en}_i\}_{i=1}^{T}$, their corresponding temporal boundaries ${\bf L}^{en} = \{({s}^{en}_i, {e}^{en}_i)\}_{i=1}^{T}$ and proposal scores ${\bf C}^{en} = \{{c}^{en}_i\}_{i=1}^{T}$. The $T$ is the number of moment proposals. Each proposal ${p}^{en}_i$  corresponds the start and end timestamps $({s}^{en}_i, {e}^{en}_i)$ and the confidence score ${c}^{en}_i \in (0, 1)$. Likewise, the suppressed branch generates ${\rm P}^{sp}$, ${\bf L}^{sp}$ and ${\bf C}^{sp}$ from the suppressed stream. Next, we can compute the enhanced score $K^{en} = \sum_{i=1}^{T}{c}^{en}_i$ and suppressed score $K^{sp} = \sum_{i=1}^{T}{c}^{sp}_i$. The intra-sample loss is given by
\begin{equation}
{\mathcal L}_{intra}  = {\rm max}(0, \ \Delta_{intra} - {K}^{en} + {K}^{sp}),
\end{equation}
where ${\mathcal L}_{intra}$ is a margin-based triplet loss and $\Delta$ is a margin which is set to 0.4. Due to the parameter sharing between two branches, the suppressed branch will select plausible negative proposals. By sufficient intra-sample confrontment, we are able to distinguish the target moment from the intractable negative moments.

Besides the intra-sample loss, we also develop a inter-sample loss by utilizing the unmatched video-sentence sample, i.e. the negative sample. Specifically, for each video $V$, we randomly select a sentence from the training set as the unmatched sentence ${\overline S}$ to form a negative sample $(V,{\overline S})$. Likewise, we can randomly choose a video to construct another negative sample $({\overline V},{S})$. Next, we apply the RTBPN to produce the enhanced scores ${\overline K}^{en}_{S}$ and ${\overline K}^{en}_{V}$ for negative samples. The inter-sample loss is given by
\begin{equation}
{\mathcal L}_{inter}  = {\rm max}(0, \ \Delta_{inter} - {K}^{en} +  {\overline K}^{en}_{S}) + {\rm max}(0, \ \Delta_{inter} - {K}^{en} +  {\overline K}^{en}_{V}),
\end{equation}
where the $\Delta_{inter}$ is set to 0.6 and ${\mathcal L}_{inter}$ encourages the enhanced scores of positive samples to be larger than negative samples.

\subsection{Language-Aware Filter}
We next introduce the language-aware filter with the scene-based cross-modal estimation.
To calculate the language-relevant score distribution over frames, we first apply a NetVLAD~\cite{arandjelovic2016netvlad} to project the textual features ${\bf Q} = \{{\bf q}_i\}_{i=1}^{n_q}$ into cluster centers. Concretely, given the trainable center vectors ${\bf C} = \{{\bf c}_j\}_{j=1}^{n_c}$ where $n_c$ is the number of centers, the NetVLAD accumulates the residuals between language features and center vectors by a soft assignment, given by
\begin{eqnarray}
& {\bf \alpha}_{i} = {\rm softmax}({\bf W}^c{\bf q}_i+ {\bf b}^c), \
{\bf u}_{j} =  \sum_{i=1}^{n_q} {\bf \alpha}_{ij}({\bf q}_i - {\bf c}_j), 
\end{eqnarray}
where ${\bf W}^{c}$ and ${\bf b}^{c}$ are projection matrix and bias. The softmax operation produces  the soft assignment coefficients ${\bf \alpha}_{i} \in \mathbb{R}^{n_c}$ corresponding to $n_c$ centers. The ${\bf u}_{j}$ is the accumulated features from ${\bf Q}$ for the $i$-th center. We can regard each center as a language scene and ${\bf u}_{j}$ is the scene-based language feature. We then calculate the cross-modal matching scores between $\{{\bf v}_i\}_{i=1}^{n_v}$ and $\{{\bf u}_j\}_{j=1}^{n_c}$ by 
\begin{eqnarray}
& \beta_{ij} = \sigma ({\bf w}^{\top}_a{\rm tanh}({\bf W}^{a}_{1} {\bf v}_{i}+ {\bf W}^{a}_{2}{\bf u}_{j}+{\bf b}^{a})),
\end{eqnarray}
where $ {\bf W}_{1}^{a}$, $ {\bf W}_{2}^{a}$ are projection matrices, ${\bf b}^{a}$ is the bias, ${\bf w}^{\top}_a$ is the row vector and $\sigma$ is the sigmoid function. The $\beta_{ij} \in (0,1)$ means the matching score of the $i$-th frame feature and $j$-th scene-based language feature. That is, scene-based method introduces an intermediately semantic space for texts and videos.

Considering a frame should be important if it is associated with any language scene, we compute the holistic score for the $i$-th frame by ${\overline \beta}_{i} = {\rm max}_j\{\beta_{ij}\}$. Then, to avoid producing a trivial score distribution, e.g. all frames are assigned to 1 or 0, we apply a max-min normalization on the distribution by 
\begin{eqnarray}
& {\widetilde \beta}_{i} = \frac{{\overline \beta}_{i} -  \min_{j}\{{\overline \beta}_{j}\}}{\max_j\{{\overline \beta}_{j}\} - \min_j\{{\overline \beta}_{j}\}}.
\end{eqnarray}
Thus, we obtain the normalized distribution $\{{\widetilde \beta}_{i}\}_{i=1}^{n_v}$ over frames, where the $i$-th value means the relevance between the $i$-th frame and language descriptions. 
Next, we apply a two-branch gate to produce the enhanced and suppressed streams, denoted by
\begin{eqnarray}
\begin{aligned}
&{\bf v}^{en}_i = {\widetilde \beta}_{i} \cdot {\bf v}_i, \ \ {\bf v}^{sp}_i = (1-{\widetilde \beta}_{i}) \cdot {\bf v}_i,
\end{aligned}
\end{eqnarray}
where the enhance stream ${\bf V}^{en} = \{{\bf v}^{en}_i\}_{i=1}^{n_v}$ highlights the critical frames and weaken unnecessary ones according to the normalized score, while the suppressed stream ${\bf V}^{sp} = \{{\bf v}^{sp}_i\}_{i=1}^{n_v}$ is the opposite.

\subsection{Sharable Two-Branch Proposal Module}
In this section, we introduce the sharable two-branch proposal module, including an enhanced branch and a suppressed branch with a consistent structure and sharable parameters. The sharing setting can make both branches produce high-quality moment proposals, avoiding the suppressed branch generating too simple negative proposals and leading to the ineffective confrontment.
Here we only present the design of the enhanced branch.

Given the enhanced stream ${\bf V}^{en} = \{{\bf v}^{en}_i\}_{i=1}^{n_v}$ and textual features ${\bf Q} = \{{\bf q}_i\}_{i=1}^{n_q}$, we first conduct a widely-used cross-modal interaction unit~\cite{zhang2019cross,chen2019localizing} to incorporate textual clues into visual features. Concretely, we perform a frame-to-word attention and aggregate the textual features for each frame, given
\begin{eqnarray}
\begin{aligned}
& \delta_{ij} = {\bf w}^{\top}_m{\rm tanh}({\bf W}^{m}_{1} {\bf v}^{en}_{i}+ {\bf W}^{m}_{2}{\bf q}_{j}+{\bf b}^{m}), \\
& {\overline \delta}_{ij} = \frac{{\rm exp}(\delta_{ij} )}{\sum_{k=1}^{n_q} {\rm exp}(\delta_{ik})},  \ {\bf s}^{en}_i = \sum_{j=1}^{n_q} {\overline \delta}_{ij} {\bf q}_{j},
\end{aligned}
\end{eqnarray}
where ${\bf s}^{en}_{i}$ is the aggregated textual representation relevant to the $i$-th frame. 
Then, the cross gate is applied to develop the visual-textual interaction, given by
\begin{eqnarray}
\begin{aligned}
&{\bf g}_i^v = \sigma({\bf W}^v{\bf v}^{en}_i + {\bf b}^v), \ {\bf g}_i^t = \sigma({\bf W}^t{\bf s}^{en}_i + {\bf b}^t),  \\ 
&{\bf \overline s}^{en}_i = {\bf s}^{en}_i \odot {\bf g}_i^v, \ {\bf \overline v}^{en}_i = {\bf v}^{en}_i \odot {\bf g}_i^t, 
\end{aligned}
\end{eqnarray}
where ${\bf g}_i^v$ is the visual gate, ${\bf g}_i^t$ is textual gate and $\odot$ is element-wise multiplication. 
After it, we concatenate ${\bf \overline v}^{en}_i$ and ${\bf \overline s}^{en}_i$ to obtain the language-aware frame feature ${\bf m}^{en}_i =  [{\bf \overline v}^{en}_i; {\bf \overline s}^{en}_i]$. 

Next, we follow the 2D temporal network~\cite{zhang2019learning} to build a 2D moment feature map and capture relationships between adjacent moments. Specifically, the 2D feature map ${\bf F} \in \mathbb{R}^{n_v \times n_v \times d_m}$ consists of three dimension: the first two dimensions represent the start and end frame indices of a moment and the third dimension is the feature dimension. The feature of a moment with temporal duration [a, b] is computed by ${\bf F}[a,b,:] = \sum_{i=a}^{b}{\bf m}^{en}_i$. Note that the location with a > b is invalid and is padded with zeros. And we also follow the sparse sampling setting in~\cite{zhang2019learning} to avoid much computational cost. That is, not all moments with a <= b are proposed if the $n_v$ is large. 
With the 2D maps, we conduct the two-layer 2D convolution with the kernel size $K$ to develop moment relationships between adjacent moments. After it, we obtain the cross-modal features $ \{{\bf f}^{en}_i\}_{i=1}^{M_{en}}$, where ${M_{en}}$ is the number of all moments in the 2D map, and compute their proposal scores $ \{{c}^{en}_i\}_{i=1}^{M_{en}}$  by
\begin{eqnarray}
& {c}^{en}_i = \sigma({\bf W}^p {\bf f}^{en}_i + {\bf b}^p).
\end{eqnarray}

Next, we employ a center-based proposal method to fiter out unnecessary moments and only retain high-quality ones as the positive moment proposals.
Concretely, we first choose the moment with the highest score ${c}^{en}_i$ as the center moment and rank the rest of moments according to the overlap with the center one. We then select top $T-1$ moments and obtain $T$ positive proposals ${\rm P}^{en} = \{{p}^{en}_i\}_{i=1}^{T}$ with proposal scores ${\bf C}^{en} = \{{c}^{en}_i\}_{i=1}^{T}$. And temporal boundaries $({s}^{en}_i, {e}^{en}_i)$ of each moment are the indices of its location in the 2D map. This method can effectively select a series of correlative moments. Likewise, the suppressed branch has the completely identical structure to generate the plausible negative proposals ${\rm P}^{sp} = \{{p}^{sp}_i\}_{i=1}^{T}$ with proposal scores ${\bf C}^{sp} = \{{c}^{sp}_i\}_{i=1}^{T}$.

\subsection{Proposal Regularization}
Next, we devise a proposal regularization strategy to inject some prior knowledge into our model, consisting of a global term and a gap term.
Due to the parameter sharing between two branches, we only apply the proposal regularization in the enhanced branch.

Specifically, considering most of moments are unaligned to the language descriptions, we first apply a global term to make the average moment score relatively low, given by 
\begin{eqnarray}
& {\mathcal L}_{global} = \frac{1}{M_{en}}\sum_{i=1}^{M_{en}} {c}^{en}_i,
\end{eqnarray}
where $M_{en}$ is the number of all moments in the 2D map.
This global term implicitly encourages the scores of unselected moments in the 2D map close to 0, while ${\mathcal L}_{intra}$ and ${\mathcal L}_{inter}$ guarantee positive proposals have high scores.

On the other hand, we further expect to identify the most accurate one as the final localization result from $T$ positive moment proposals, thus it is crucial to enlarge the score gaps between these proposals to make them distinguishable. We perform ${\rm softmax}$ on positive proposal scores and then employ the gap term ${\mathcal L}_{gap}$  by
\begin{eqnarray}
& {\overline c}^{en}_i = \frac{{\rm exp}({c}^{en}_i)}{\sum_{i=1}^{T}{\rm exp}({c}^{en}_i)}, \ \ {\mathcal L}_{gap}  = -\sum_{i=1}^{T} {\overline c}^{en}_i  {\rm log}({\overline c}^{en}_i), 
\end{eqnarray}
where $T$ is the number of positive proposals rather than the number $M_{en}$ of all proposals.
When the ${\mathcal L}_{gap}$ decreases, the score distribution will become more diverse, i.e. it implicitly encourages to enlarge the score gaps between positive moment proposals.

\subsection{Training and Inference}
Based on the aforementioned model design, we apply a multi-task loss to train our RTBPN in an end-to-end manner, given by
\begin{eqnarray}
\begin{aligned}
{\mathcal L}_{RTBPN} = {\lambda}_1 {\mathcal L}_{intra} +  {\lambda}_2 {\mathcal L}_{inter} + {\lambda}_3 {\mathcal L}_{global} + {\lambda}_4 {\mathcal L}_{gap},
\end{aligned}
\end{eqnarray}
where ${\lambda}_*$ are the hyper-parameters to control the balance of losses. 
% Actually, we only need to keep these loss values at the same magnitude and the elaborate adjustment is unnecessary.

During inference, we can directly select the moment ${p}^{en}_i$  with the highest proposal score ${c}^{en}_i$ from the enhanced branch.

\section{Experiments}

\subsection{Datasets}
We conduct extensive experiments on three public datasets.

\textbf{Charades-STA~\cite{gao2017tall}:} The dataset is built on the original Charades dataset~\cite{sigurdssonhollywood}, where Gao et al. apply a semi-automatic way to generate the language descriptions for temporal moments. This dataset contains 9,848 videos of indoor activities and their average duration is 29.8 seconds. 
The dataset contains 12,408 sentence-moment pairs for training and 3,720 pairs for testing.

\textbf{ActivityCaption~\cite{caba2015activitynet}:} The dataset contains 19,209 videos with diverse contents and their average duration is about 2 minutes. Following the standard split in~\cite{zhang2019cross,zhang2019learning}, there are 37,417, 17,505 and 17,031 sentence-moment pairs used for training, validation and testing, respectively. This is the largest dataset currently.

\textbf{DiDeMo~\cite{hendricks2017localizing}:} The dataset consists of 10,464 videos and the duration of each video is 25-30 seconds. It contains 33,005 sentence-moment pairs for training, 4,180 for validation and 4,021 for testing. Especially, each video in DiDeMo is divided into six five-second clips and the target moment contains one or more consecutive clips. Thus, there are only 21 moment candidates while Charades-STA and ActivityCaption allow arbitrary temporal boundaries.

\subsection{Evaluation Criteria}
Following the widely-used setting~\cite{gao2017tall,hendricks2017localizing}, we apply {\bf R@n,IoU=m} as the criteria for Charades-STA and ActivityCaption and use {\bf Rank@1}, {\bf Rank@5} and {\bf mIoU} as the criteria for DiDeMo. 
Concretely, we first calculate the IoU between the predicted moments and ground truth, and {\bf R@n,IoU=m} means the percentage of at least one of the top-n moments having the IoU $>$ m. The {\bf mIoU} is the average IoU of the top-1 moment over all testing samples. 
And for DiDeMo, due to only 21 moment candidates, {\bf Rank@1} or {\bf Rank@5} is the percentage of samples which ground truth moment is ranked as top-1 or among top-5.

\begin{table}[t]
    \centering
    \caption{Performance Evaluation Results on Charades-STA 
    ($n \in \{1,5\}$ and  $m \in \{0.3,0.5,0.7\}$). }
    \label{table:charades}
    \scalebox{0.8}{
        \begin{tabular}{c|ccc|ccc}
            \hline
            \multirow{2}{*}{Method} & \multicolumn{3}{c|}{R@1} & \multicolumn{3}{c}{R@5} \\
                &  IoU=0.3&   IoU=0.5 &  IoU=0.7 & IoU=0.3&   IoU=0.5 &  IoU=0.7\\
            \hline
            \hline
            \multicolumn{7}{c}{fully-supervised methods}  \\
            \hline
                VSA-RNN~\cite{gao2017tall}&-&10.50&4.32
                       &-&48.43&20.21\\
                VSA-STV~\cite{gao2017tall}&-&16.91&5.81
                       &-&53.89&23.58\\
                CTRL~\cite{gao2017tall}&-&23.63&8.89
                    &-&58.92&29.52\\
                QSPN~\cite{xu2019multilevel} &54.70&35.60&15.80
                    &95.60&79.40&45.40\\
                2D-TAN~\cite{zhang2019learning}&-&39.81&23.25
                    &-&79.33&52.15\\            
            \hline
            \hline
            \multicolumn{7}{c}{weakly-supervised methods}  \\
            \hline
                TGA~\cite{mithun2019weakly}&32.14&19.94&8.84
                   &86.58&65.52&33.51\\
                CTF~\cite{chen2020look}&39.80&27.30&12.90
                      &-&-&-\\
                SCN~\cite{lin2019weakly}&{42.96}&{23.58}&{9.97}
                   &{95.56}&{71.80}&{38.87}\\
                RTBPN~(our) &{\bf 60.04}&{\bf 32.36}&{\bf 13.24}
                   &{\bf 97.48}&{\bf 71.85}&{\bf 41.18}\\
            \hline
        \end{tabular}
    }
\end{table}

\subsection{Implementation Details}
We next introduce the implementation details of our RTBPN model.

\textbf{Data Preprocessing.} For a fair comparison, we apply the same visual features as previous methods~\cite{gao2017tall,hendricks2017localizing,zhang2019cross}, that is, C3D features for Charades-STA and ActivityCaption and VGG16 and optical flow features for DiDeMo. We then shorten the feature sequence using temporal mean pooling with the stride 4 and 8 for Charades-STA and ActivityCaption, respectively. And for DiDeMo, we compute the average feature for each fixed five-second clips as in~\cite{hendricks2017localizing}.
As for sentence queries, we extract 300-d word embeddings by the pre-trained Glove embedding~\cite{pennington2014glove} for each word token.

\textbf{Model Setting.} In the center-based proposal method, the positive/negative proposal number $T$ is set to 48 for Charades-STA and ActivityCaption and 6 for DiDeMo. During 2D feature map construction, we fill all locations [a, b] if a <= b for DiDeMo. But for Charades-STA, we add another limitation $(b-a) \mod 2 = 1$. And for ActivityCaption, we only fill the location [a, b] if $(b-a) \mod 8 = 0$. The
sparse sampling avoids much computational cost. 
We set the convolution kernel size $K$ to 3, 9 and 3 for Charades-STA, ActivityCaption and DiDeMo, respectively.
Besides, the dimension of almost parameter matrices and bias in our model to 256, including the ${\bf W}^{c}$, ${\bf b}^{c}$ in the NetVLAD, ${\bf W}^{m}_1$, ${\bf W}^{m}_2$ and ${\bf b}^{m}$ in the frame-to-word attention and so on.  We set the dimension of the hidden state of each direction in the Bi-GRU networks to 128. And the dimension of trainable center vectors is 256.
During training, we set $\lambda_1$,  $\lambda_2$,  $\lambda_3$, $\lambda_4$ to 0.1, 1, 0.01 and 0.01, respectively. And we use an Adam optimizer~\cite{duchi2011adaptive} with the initial learning rate 0.001 and batch size 64. During inference, we apply the non-maximum suppression (NMS) with a threshold 0.55 while we need to select multiple moments.

\begin{table}[t]
    \centering
    \caption{Performance Evaluation Results on ActivityCaption 
    ($n \in \{1,5\}$ and  $m \in \{0.1,0.3,0.5\}$). }
    \label{table:activity}
    \scalebox{0.8}{
        \begin{tabular}{c|ccc|ccc}
            \hline
            \multirow{2}{*}{Method} & \multicolumn{3}{c|}{R@1} & \multicolumn{3}{c}{R@5} \\
                &  IoU=0.1&   IoU=0.3 &  IoU=0.5 & IoU=0.1&   IoU=0.3 &  IoU=0.5 \\
            \hline
            \hline
            \multicolumn{7}{c}{fully-supervised methods}  \\
            \hline
                TGN~\cite{chen2018temporally}&-&43.81&27.93
                    &-&54.56&44.20\\  
                QSPN~\cite{xu2019multilevel} &-&45.30&27.70
                    &-&75.70&59.20\\
                2D-TAN~\cite{zhang2019learning}&-&59.45&44.51
                    &-&85.53&77.13\\  

            \hline
            \hline
            \multicolumn{7}{c}{weakly-supervised methods}  \\
            \hline
                WS-DEC~\cite{duan2018weakly}&62.71&41.98&23.34
                      &-&-&-\\
                WSLLN~\cite{gao2019wslln} &{\bf 75.40}&42.80&22.70
                      &-&-&-\\
                CTF~\cite{chen2020look}&{74.20}&44.30&23.60
                      &-&-&-\\
                SCN~\cite{lin2019weakly}&{71.48}&{47.23}&{29.22}
                   &{90.88}&{71.45}&{55.69}\\
                RTBPN~(our)&{73.73}&{\bf 49.77}&{\bf 29.63}
                   &{\bf 93.89}&{\bf 79.89}&{\bf 60.56}\\
            \hline
        \end{tabular}
    }
\end{table}

\begin{table}[t]
    \centering
    \caption{Performance Evaluation Results on DiDeMo. }
    \label{table:didemo}
        \begin{tabular}{c|c|ccc}
            \hline
            {Method}&  Input&   Rank@1 &  Rank@5 &  mIoU\\
            \hline
            \hline
            \multicolumn{5}{c}{fully-supervised methods}  \\
            \hline
                MCN~\cite{hendricks2017localizing}&RGB&13.10&44.82&25.13\\
                TGN~\cite{chen2018temporally}&RGB&24.28&71.43&38.62\\
                \hline
                MCN~\cite{hendricks2017localizing}&Flow&18.35&56.25&31.46\\
                TGN~\cite{chen2018temporally}&Flow&27.52&76.94&42.84\\
                \hline
                MCN~\cite{hendricks2017localizing}&RGB+Flow&28.10&78.21&41.08\\
                TGN~\cite{chen2018temporally}&RGB+Flow&28.23&79.26&42.97\\
            \hline
            \hline
            \multicolumn{5}{c}{weakly-supervised methods}  \\
            \hline
                WSLLN~\cite{gao2019wslln} &RGB&19.40&53.10&25.40\\
                RTBPN~(our) &RGB&{\bf 20.38}&{\bf 55.88}&{\bf 26.53}\\
                \hline
                WSLLN~\cite{gao2019wslln} &Flow&18.40&54.40&27.40\\
                RTBPN~(our) &Flow&{\bf 20.52}&{\bf 57.72}&{\bf 30.54}\\
                \hline
                TGA~\cite{mithun2019weakly}&RGB+Flow&12.19&39.74&24.92\\
                RTBPN~(our) &RGB+Flow&{\bf 20.79}&{\bf 60.26}&{\bf 29.81}\\
            \hline
        \end{tabular}
\end{table}

\begin{table*}[t]
    \centering
    \caption{Ablation results about the two-branch architecture, filter details and center-based proposal method.}
    \label{table:design}
    \scalebox{0.9}{
        \begin{tabular}{c|ccc|ccc|ccc|ccc}
            \hline
            \multirow{3}{*}{Method} & \multicolumn{6}{c|}{Charades-STA} & \multicolumn{6}{c}{ActivityCaption} \\
            \cline{2-13}
             & \multicolumn{3}{c|}{R@1} & \multicolumn{3}{c|}{R@5} & 
            \multicolumn{3}{c|}{R@1} & \multicolumn{3}{c}{R@5} \\
                &  IoU=0.3&   IoU=0.5 &  IoU=0.7 & IoU=0.3&   IoU=0.5 &  IoU=0.7 &
                   IoU=0.1&   IoU=0.3 &  IoU=0.5 & IoU=0.1&   IoU=0.3 &  IoU=0.5 \\
            \hline
            \hline
            \multicolumn{13}{c}{The Two-Branch Architecture} \\
            \hline
            w/o. filter  &56.43&29.14&11.40
                     &94.86&67.25&37.59
                     &73.54& 43.55& 26.67
                     &89.79&73.14&57.92\\
            w/o. parameter sharing   &{32.62}&13.87&4.55
                      &80.43&47.06&19.28
                      &{\bf 80.47}&48.35&22.92
                      &90.27&75.11&57.03\\
            full model &{\bf 60.04}&{\bf 32.36}&{\bf 13.24}
                   &{\bf 97.48}&{\bf 71.85}&{\bf 41.18}
                   &{73.73}&{\bf 49.77}&{\bf 29.63}
                   &{\bf 93.89}&{\bf 79.89}&{\bf 60.56}\\
            \hline
            \hline
            \multicolumn{13}{c}{The Filter Design} \\
            \hline
            visual-only scoring   &{57.85}&{30.59}&{12.89}
                   &{95.78}&{68.75}&{40.54}
                   &71.82&45.69&27.87
                    &90.52&76.03&58.87\\
            w/o. NetVALD&{58.61}&31.92&13.14
                      &96.26&70.84&40.70
                      &{72.32}&45.15&28.08
                      &91.41&77.75&59.91\\
            full model &{\bf 60.04}&{\bf 32.36}&{\bf 13.24}
                   &{\bf 97.48}&{\bf 71.85}&{\bf 41.18}
                   &{\bf 73.73}&{\bf 49.77}&{\bf 29.63}
                   &{\bf 93.89}&{\bf 79.89}&{\bf 60.56}\\
            \hline
            \hline
            \multicolumn{13}{c}{The Proposal Method} \\
            \hline
            all-proposal&{57.92}&{30.94}&{12.16}
                 &{95.59}&{68.21}&{38.84}
                  &{\bf 82.61}&48.02&21.21
                  &90.37&73.09&55.02\\
            top-k proposal&{58.61}&{31.16}&{12.63}
                 &{95.38}&{69.70}&{39.55}
                 &71.85&47.08&28.25
                 &92.82&77.63&59.89\\

            full model (center-based) &{\bf 60.04}&{\bf 32.36}&{\bf 13.24}
                   &{\bf 97.48}&{\bf 71.85}&{\bf 41.18}
                   &{73.73}&{\bf 49.77}&{\bf 29.63}
                   &{\bf 93.89}&{\bf 79.89}&{\bf 60.56}\\
            \hline

        \end{tabular}
    }
\end{table*}

\subsection{Comparison to State-of-the-Art Methods} 
We compare our RTBPN method with existing state-of-the-art methods, including the supervised and weakly-supervised approaches.

\textbf{Supervised Method:} Early approaches VSA-RNN~\cite{gao2017tall}, VSA-STV~\cite{gao2017tall}, CTRL~\cite{gao2017tall} and MCN~\cite{hendricks2017localizing} projects the visual features of candidate moments and textual features into a common space for correlation estimation. From a holistic view, TGN~\cite{chen2018temporally} develops the frame-by-word interaction by RNN. And QSPN~\cite{xu2019multilevel} integrates vision and language features early and re-generate descriptions as an auxiliary task.
Further, 2D-TAN~\cite{zhang2019learning} captures the temporal relations between adjacent moments by the 2D moment map.

\textbf{Weakly-Supervised Method:} WS-DEC~\cite{duan2018weakly} regards weakly-supervised moment retrieval and dense video captioning as the dual problems. Under the MIL framework, TGA~\cite{mithun2019weakly} utilizes the text-guided attention weights to detect the target moment, WSLLN~\cite{gao2019wslln} simultaneously apply the alignment and detection module to boost the performance, and CTF~\cite{chen2020look} detects the moment in a two-stage coarse-to-fine manner.
Different from MIL-based methods, SCN~\cite{lin2019weakly} ranks moment proposals by a language reconstruction reward.

The overall evaluation results on three large-scale datasets are presented in Table~\ref{table:charades}, Table~\ref{table:activity} and Table~\ref{table:didemo}, where we set $n \in \{1,5\}, m \in \{0.3,0.5,0.7\}$ for Charades-STA and $n \in \{1,5\}, m \in \{0.1,0.3,0.5\}$ for ActivityCaption. The results reveal some interacting points:
\begin{itemize}
    \item On almost all criteria of three datasets, our RTBPN method achieves the best weakly-supervised performance, especially on Charades-STA. This fact verifies the effectiveness of our two-branch framework with the regularization strategy.
    \item The reconstruction-based method SCN outperforms MIL-based methods TGA, CTF and WSLLN on Charades-STA and ActivityCaption, but our RTBPN achieves a better performance than SCN, demonstrating our RTBPN with the intra-sample confrontment can effectively discover the plausible negative samples and improve the accuracy.
    \item On the DiDeMo dataset, our RTBPN outperforms the state-of-the-art baselines using RGB, Flow and two-stream features. This fact suggests our method is robust for diverse features.
    \item Our RTBPN outperforms the early supervised approaches VSA-RNN, VSA-STV, CTRL and obtains the results comparable to other methods TGN, QSPN and MCN, which indicates even under the weakly-supervised setting, our RTBPN can still develop the sufficient visual-language interacting and retrieve the accurate moment.
\end{itemize}

\begin{figure}[t]
    \centering
    \subfigure[Charades-STA]{
        \includegraphics[width=0.47\columnwidth]{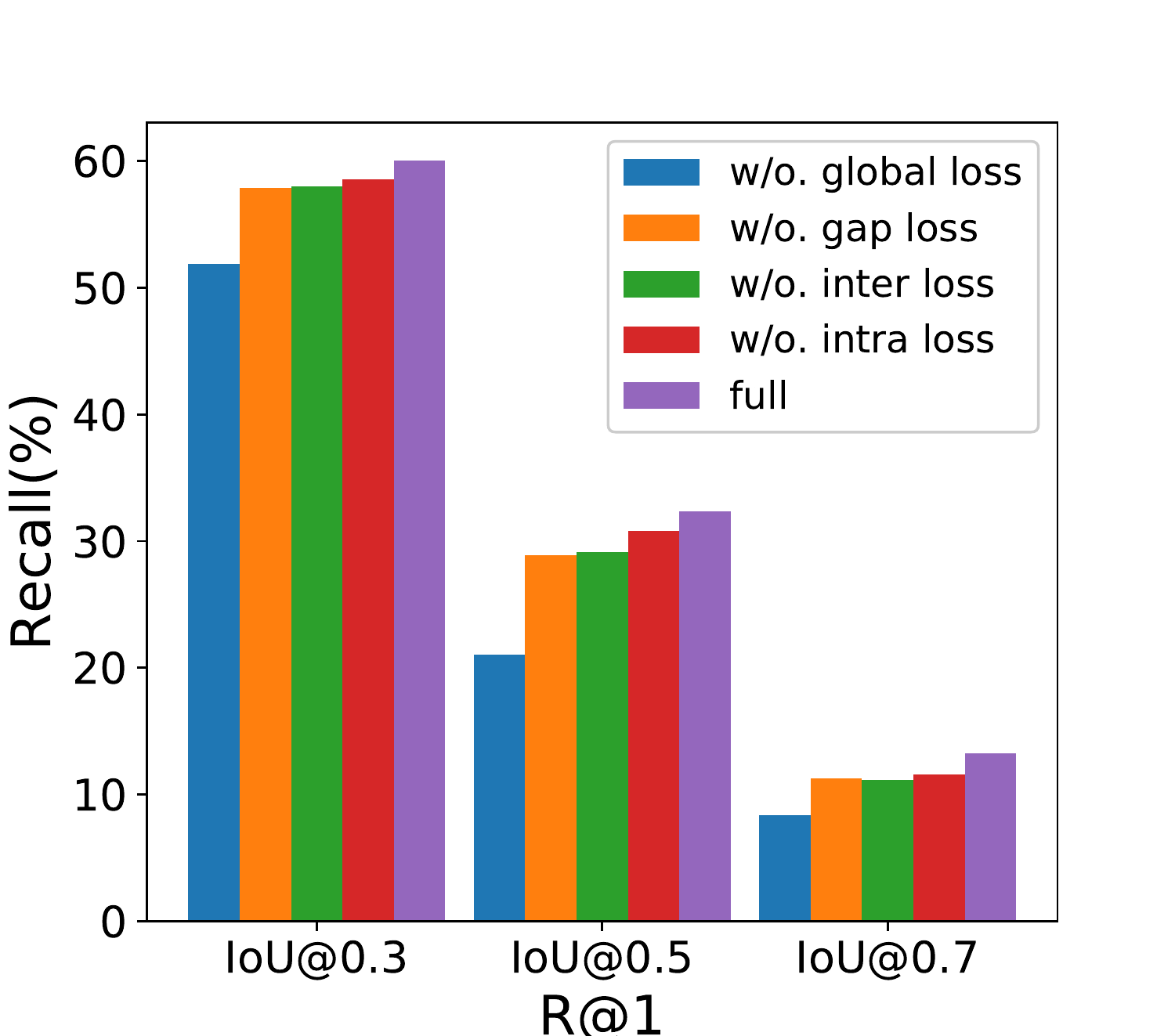}
    }
    \subfigure[ActivityCaption]{
        \includegraphics[width=0.47\columnwidth]{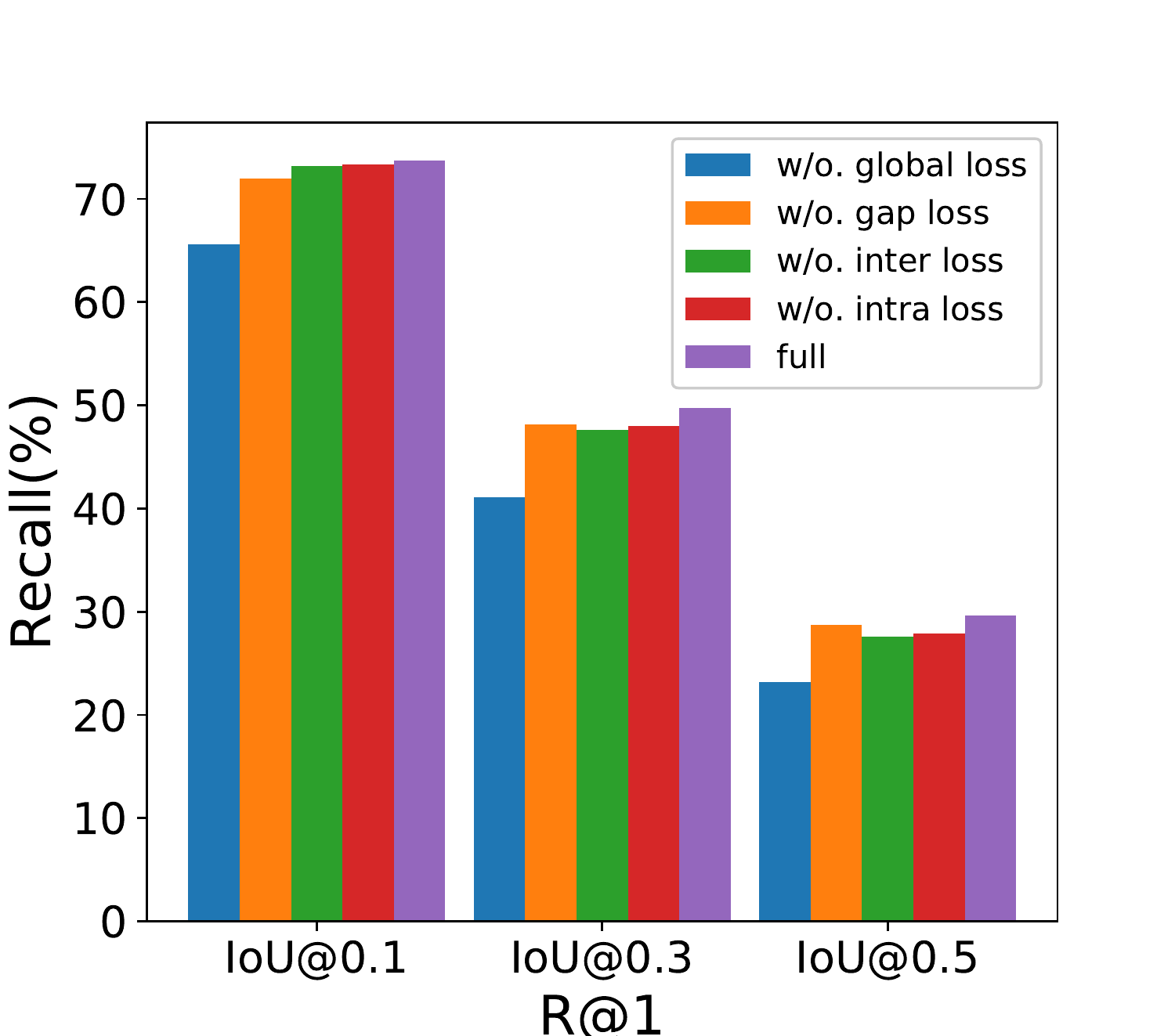}
    }
    \caption{Ablation Results of the Multi-Task Losses.}
    \label{fig:loss} 
\end{figure}

\begin{figure}[t]
    \centering
    \subfigure[Charades-STA]{
        \includegraphics[width=0.47\columnwidth]{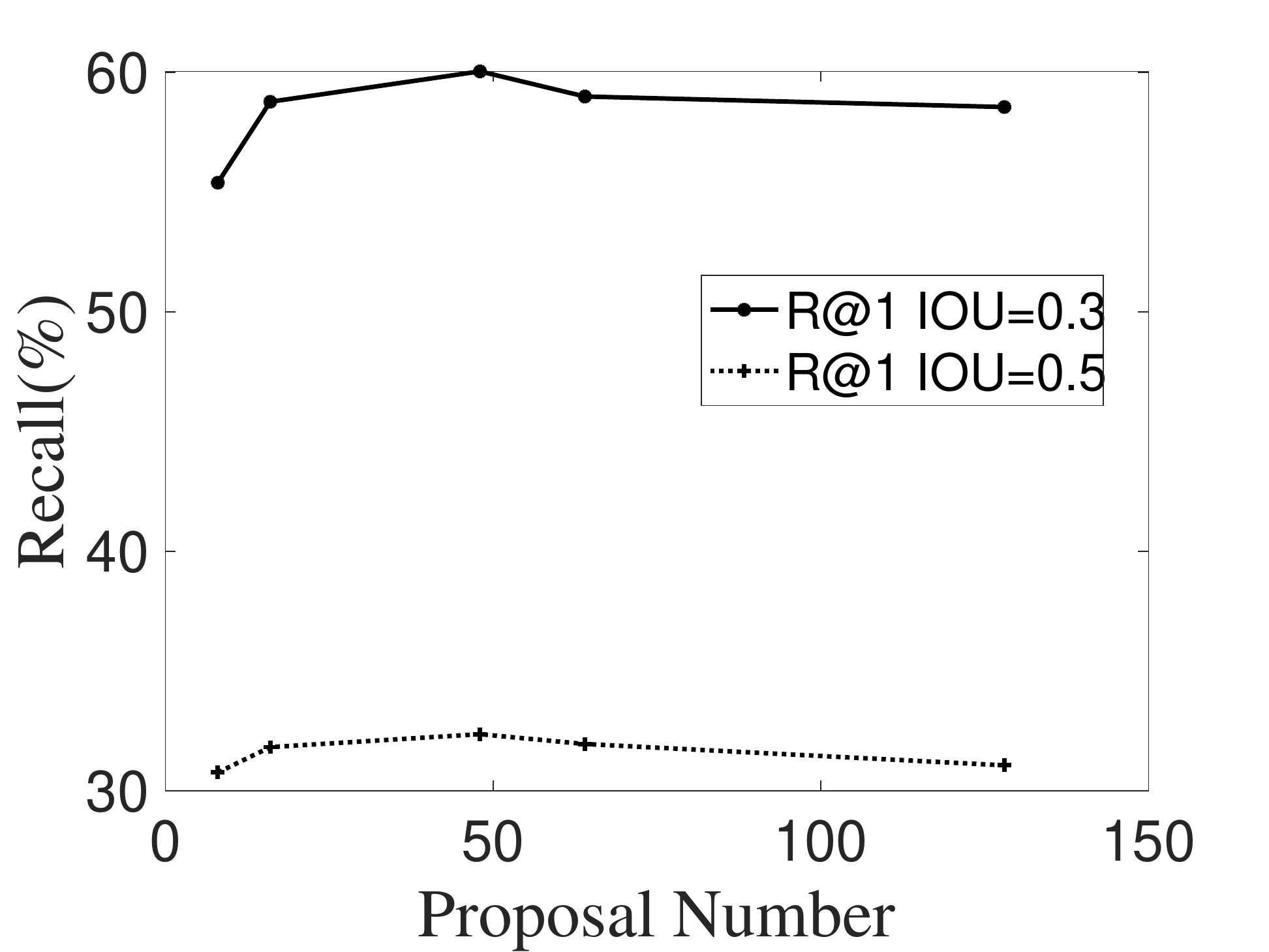}
    }
    \subfigure[ActivityCaption]{
        \includegraphics[width=0.47\columnwidth]{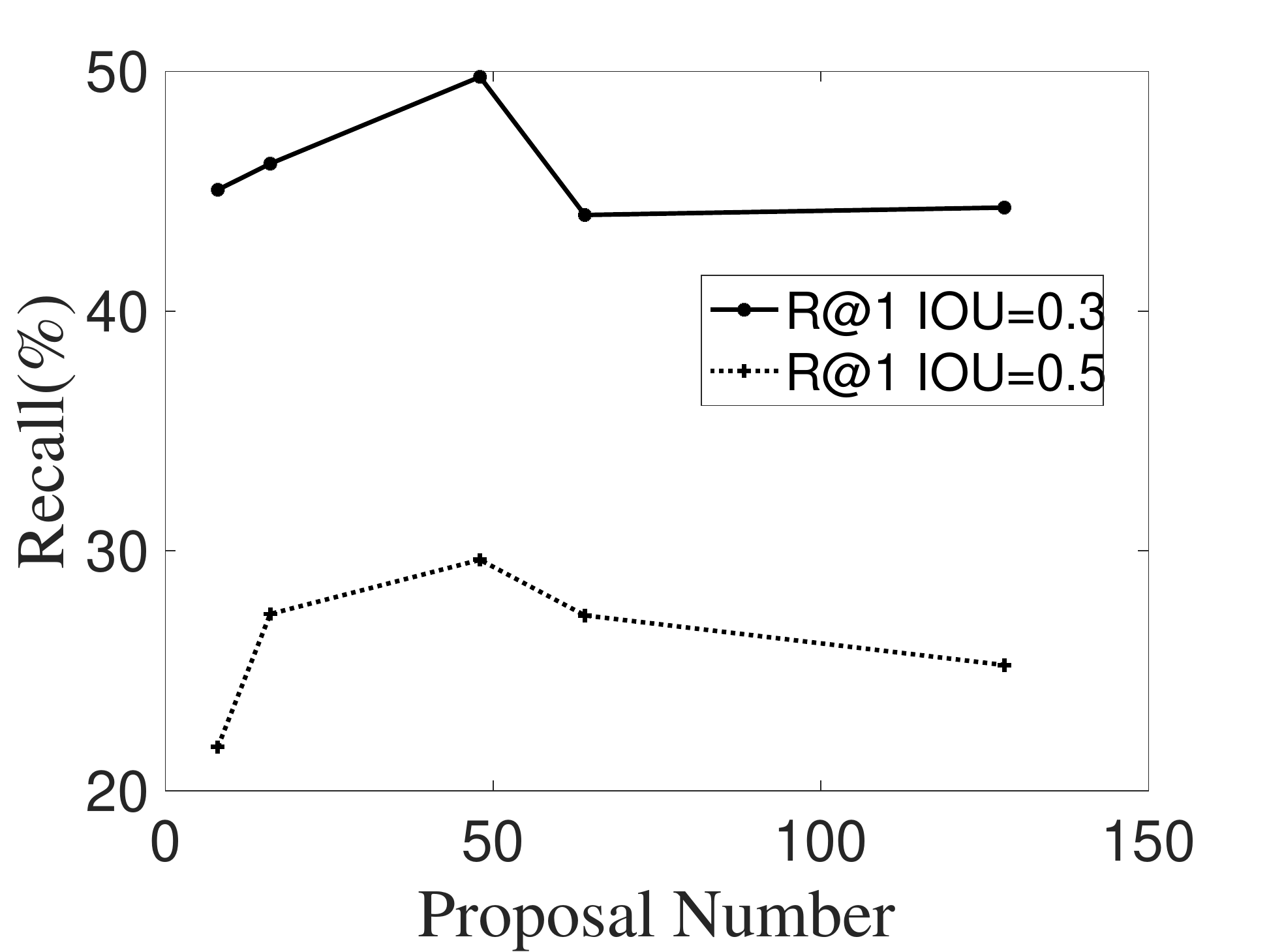}
    }
    \caption{Effect of the Proposal Number 
    on Charades-STA and ActivityCaption Datasets.}
    \label{fig:hyper} 
\end{figure}

\subsection{Ablation Study}
In this section, we conduct the ablation study for the multi-task loss and the concrete design of our model. 

\subsubsection{Ablation Study for the Multi-Task Loss}
We discard one loss from the multi-task loss at a time to generate an ablation model, including \textbf{w/o. intra loss}, \textbf{w/o. inter loss} and so on. The ablation results are shown in Figure~\ref{fig:loss}. 
We can find the full model outperforms all ablation models on two datasets, which demonstrates the intra-sample and inter-sample losses can effectively offer the supervision signals, and the regularized global and gap losses can improve the model performance. The model~(w/o. inter loss) and model~(w/o. intra loss) have close performance, suggesting intra-sample and inter-sample confrontments are equally important for weakly-supervised moment retrieval. Moreover, the model~(w/o. global loss) achieves the worst accuracy, which shows filtering out irrelevant moments is crucial to model training.

\subsubsection{Ablation Study for the Model Design}
We next verify the effectiveness of our model design, including the two-branch architecture, filter designs and center-based proposal method. Note that the cross-modal interesting unit~\cite{zhang2019cross} and 2D temporal map~\cite{zhang2019learning} are mature techniques that do not need further ablation.
\begin{itemize}
    \item \textbf{Two-Branch Architecture.} We remove the crucial filter and only retain a branch to perform the conventional MIL-based training without the intra-sample loss as \textbf{w/o. filter}. We then keep the entire framework but discard the parameter sharing between two branches as \textbf{w/o. parameter sharing}.
    \item \textbf{Filter Design.} We discard the cross-modal estimation and generate the score distribution by only frame features as \textbf{w/o. visual-only scoring}. And we remove the NetVALD and directly apply the textual features during the cross-modal estimation as \textbf{w/o. NetVALD}.
    \item \textbf{Proposal Method.} During moment proposal generation in two branches, we discard the center-based proposal and sample all candidate moments as \textbf{all-proposal}. And we replace the center-based proposal method with a top-k proposal method as \textbf{top-k proposal}, where we directly select $T$ moments with the high proposal scores.
\end{itemize}
The ablation results on ActivityCaption and Charades-STA datasets are reported in Table~\ref{table:design} and we can find some interesting points:
\begin{itemize}
    \item The model~(w/o. filter) and model~(w/o. parameter sharing) have severe performance degradation than the full model. This fact demonstrates that the two-branch architecture with the language-aware filter can develop the intra-sample confrontment and boost the model performance, and the parameter sharing is crucial to make two branches generate high-quality proposals for sufficient confrontment.
    \item The full model achieves better results than model~(visual-only scoring) and model~(w/o. NetVALD). It suggests that the cross-modal estimation with language information can generate a more reasonable score distribution than visual-only scoring. And the NetVALD can further enhance the cross-modal estimation by introducing an intermediately semantic space for texts and videos.
    \item As for the proposal method, the model with the center-based strategy outperforms the model~(all-proposal) and model~(top-k proposal), which proves our center-based proposal method can discover a series of correlative moments for MIL-based intra-sample and inter-sample training.
    \item Actually, some ablation models, e.g. model~(visual-only scoring) and model~(top-k proposal), still yield better performance than state-of-the-art baselines, validating our RTBPN network is robust and does not depend on a key component.
\end{itemize}

\begin{figure}[t]
    \centering
    \includegraphics[width=1.0\columnwidth]{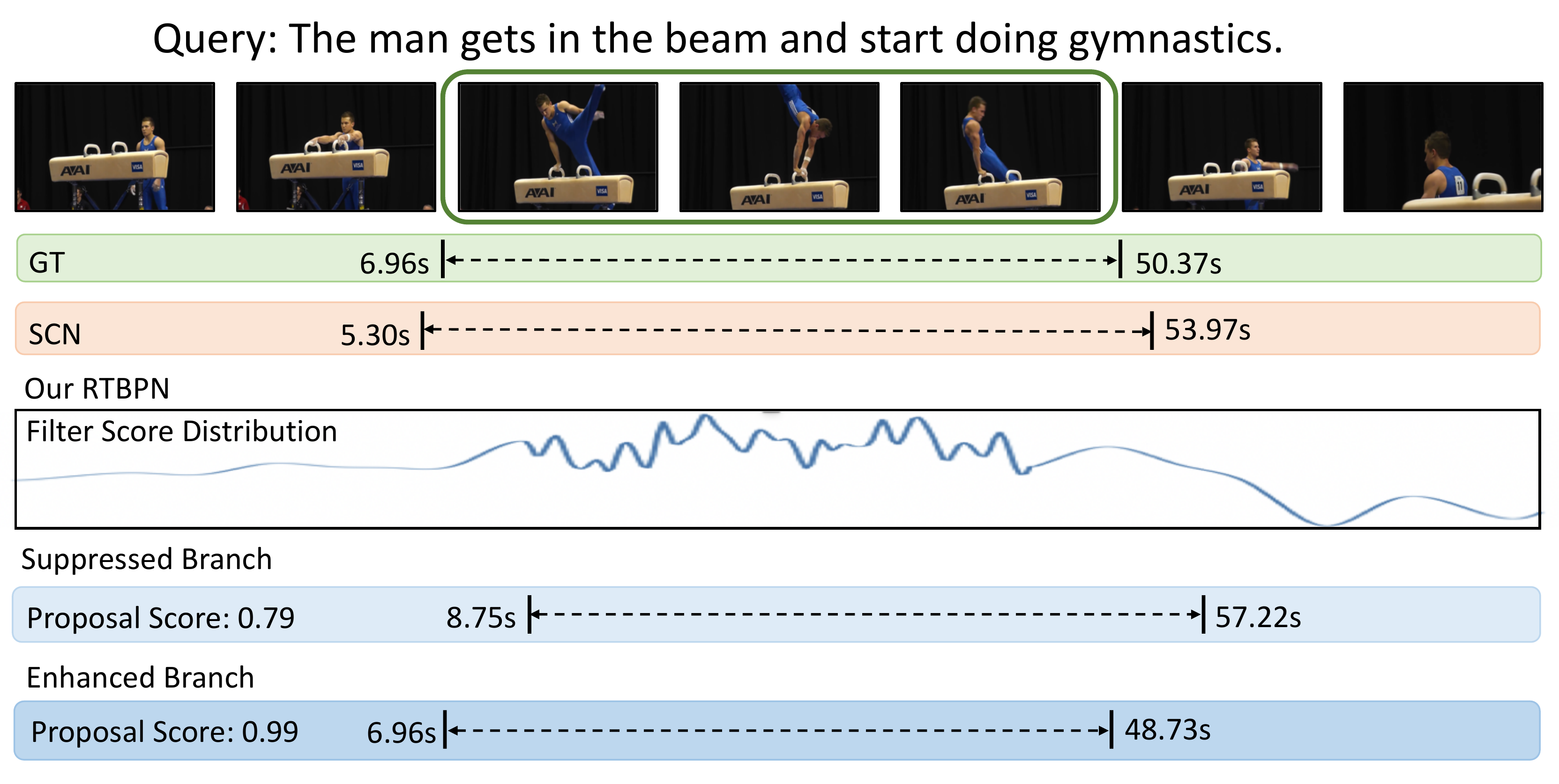} 
    \includegraphics[width=1.0\columnwidth]{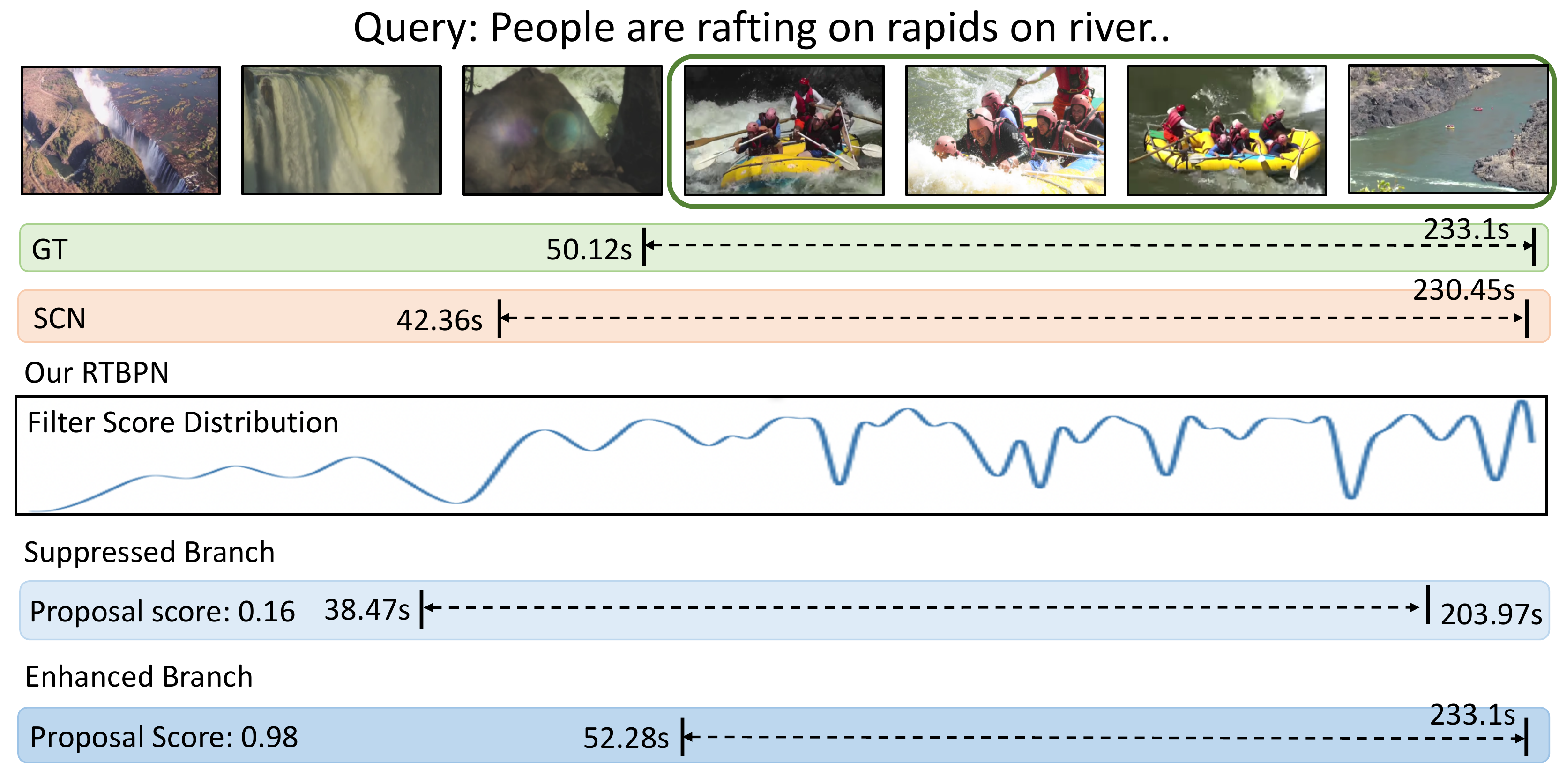} 
    \caption{Qualitative Examples on the Charades-STA and ActivityCaption datasets}
    \label{fig:example_act}
\end{figure}

\subsection{Hyper-Parameters Analysis}
In our RTBPN model, the number of selected positive/negative proposal number $T$ is an important hyper-parameter. Therefore, we further explore its effect by varying the proposal number. Specifically, we set $T$ to 8, 16, 48, 64, 128 on ActivityCaption and Charades-STA datasets and report the experiment results in Figure~\ref{fig:hyper}, where we select "R@1, IoU=0.3" and "R@1, IoU=0.5" as evaluation criteria. We note that the model achieves the best performance on both datasets when the number is set to 48. Because too many proposals will introduce irrelevant moments in the model training and affect the model performance. And too few proposals may miss the crucial moments and fail to develop sufficient confrontment, leading to poor performance. 
Moreover, the trends of the effect of proposal number $T$ on two datasets are similar, which demonstrates this hyper-parameter is insensitive to different datasets.

\subsection{Qualitative Analysis}
To qualitatively validate the effectiveness of our RTBPN method, we display two typical examples on  ActivityCaption and Charades-STA in Figure~\ref{fig:example_act}, where we show the score distribution from the language-aware filter, the retrieval results from the enhanced branch and suppressed branch and the result of the SCN baseline. 

By intuitive comparison, we find that our RTBPN method can retrieve a more accurate moment from the enhanced branch than SCN, qualitatively verifying the effectiveness of our method.
And we can observe that the filter gives higher scores to the language-relevant frames than unnecessary ones. Based on the reasonable score distribution, the enhanced branch can localize the precise moment while the suppressed branch can only retrieve the relevant but not accurate moment as the plausible negative proposal.

\section{Conclusion}
In this paper, we propose a novel regularized two-branch proposal network for weakly-supervised video moment retrieval. We devise a language-aware filter to generate the enhanced and suppressed video streams, and then design the sharable two-branch proposal module to generate positive proposals from the enhanced stream and plausible negative proposals from the suppressed one.
Further, we design the proposal regularization to improve the model performance. The extensive experiments show the effectiveness of our RTBPN method.

%%
%% The acknowledgments section is defined using the "acks" environment
%% (and NOT an unnumbered section). This ensures the proper
%% identification of the section in the article metadata, and the
%% consistent spelling of the heading.
\begin{acks}
This work is supported by the National Key R\&D Program of China under Grant No. 2018AAA0100603, Zhejiang Natural Science Foundation LR19F020006 and the National Natural Science Foundation of China under Grant No.61836002, No.U1611461 and No.61751209.
This research is supported by the Fundamental Research Funds for the Central Universities 2020QNA5024.
\end{acks}

%%
%% The next two lines define the bibliography style to be used, and
%% the bibliography file.
\clearpage
\bibliographystyle{ACM-Reference-Format}
\balance
\bibliography{all}

%%% -*-BibTeX-*-
%%% Do NOT edit. File created by BibTeX with style
%%% ACM-Reference-Format-Journals [18-Jan-2012].

\begin{thebibliography}{44}

%%% ====================================================================
%%% NOTE TO THE USER: you can override these defaults by providing
%%% customized versions of any of these macros before the \bibliography
%%% command.  Each of them MUST provide its own final punctuation,
%%% except for \shownote{}, \showDOI{}, and \showURL{}.  The latter two
%%% do not use final punctuation, in order to avoid confusing it with
%%% the Web address.
%%%
%%% To suppress output of a particular field, define its macro to expand
%%% to an empty string, or better, \unskip, like this:
%%%
%%% \newcommand{\showDOI}[1]{\unskip}   % LaTeX syntax
%%%
%%% \def \showDOI #1{\unskip}           % plain TeX syntax
%%%
%%% ====================================================================

\ifx \showCODEN    \undefined \def \showCODEN     #1{\unskip}     \fi
\ifx \showDOI      \undefined \def \showDOI       #1{#1}\fi
\ifx \showISBNx    \undefined \def \showISBNx     #1{\unskip}     \fi
\ifx \showISBNxiii \undefined \def \showISBNxiii  #1{\unskip}     \fi
\ifx \showISSN     \undefined \def \showISSN      #1{\unskip}     \fi
\ifx \showLCCN     \undefined \def \showLCCN      #1{\unskip}     \fi
\ifx \shownote     \undefined \def \shownote      #1{#1}          \fi
\ifx \showarticletitle \undefined \def \showarticletitle #1{#1}   \fi
\ifx \showURL      \undefined \def \showURL       {\relax}        \fi
% The following commands are used for tagged output and should be
% invisible to TeX
\providecommand\bibfield[2]{#2}
\providecommand\bibinfo[2]{#2}
\providecommand\natexlab[1]{#1}
\providecommand\showeprint[2][]{arXiv:#2}

\bibitem[\protect\citeauthoryear{Arandjelovic, Gronat, Torii, Pajdla, and
  Sivic}{Arandjelovic et~al\mbox{.}}{2016}]%
        {arandjelovic2016netvlad}
\bibfield{author}{\bibinfo{person}{Relja Arandjelovic}, \bibinfo{person}{Petr
  Gronat}, \bibinfo{person}{Akihiko Torii}, \bibinfo{person}{Tomas Pajdla},
  {and} \bibinfo{person}{Josef Sivic}.} \bibinfo{year}{2016}\natexlab{}.
\newblock \showarticletitle{NetVLAD: CNN architecture for weakly supervised
  place recognition}. In \bibinfo{booktitle}{\emph{Proceedings of the IEEE
  Conference on Computer Vision and Pattern Recognition}}.
  \bibinfo{pages}{5297--5307}.
\newblock


\bibitem[\protect\citeauthoryear{Caba~Heilbron, Escorcia, Ghanem, and
  Carlos~Niebles}{Caba~Heilbron et~al\mbox{.}}{2015}]%
        {caba2015activitynet}
\bibfield{author}{\bibinfo{person}{Fabian Caba~Heilbron},
  \bibinfo{person}{Victor Escorcia}, \bibinfo{person}{Bernard Ghanem}, {and}
  \bibinfo{person}{Juan Carlos~Niebles}.} \bibinfo{year}{2015}\natexlab{}.
\newblock \showarticletitle{Activitynet: A large-scale video benchmark for
  human activity understanding}. In \bibinfo{booktitle}{\emph{Proceedings of
  the IEEE Conference on Computer Vision and Pattern Recognition}}.
  \bibinfo{pages}{961--970}.
\newblock


\bibitem[\protect\citeauthoryear{Chao, Vijayanarasimhan, Seybold, Ross, Deng,
  and Sukthankar}{Chao et~al\mbox{.}}{2018}]%
        {chao2018rethinking}
\bibfield{author}{\bibinfo{person}{Yu-Wei Chao}, \bibinfo{person}{Sudheendra
  Vijayanarasimhan}, \bibinfo{person}{Bryan Seybold}, \bibinfo{person}{David~A
  Ross}, \bibinfo{person}{Jia Deng}, {and} \bibinfo{person}{Rahul Sukthankar}.}
  \bibinfo{year}{2018}\natexlab{}.
\newblock \showarticletitle{Rethinking the Faster R-CNN Architecture for
  Temporal Action Localization}. In \bibinfo{booktitle}{\emph{Proceedings of
  the IEEE Conference on Computer Vision and Pattern Recognition}}.
  \bibinfo{pages}{1130--1139}.
\newblock


\bibitem[\protect\citeauthoryear{Chen, Chen, Ma, Jie, and Chua}{Chen
  et~al\mbox{.}}{2018}]%
        {chen2018temporally}
\bibfield{author}{\bibinfo{person}{Jingyuan Chen}, \bibinfo{person}{Xinpeng
  Chen}, \bibinfo{person}{Lin Ma}, \bibinfo{person}{Zequn Jie}, {and}
  \bibinfo{person}{Tat-Seng Chua}.} \bibinfo{year}{2018}\natexlab{}.
\newblock \showarticletitle{Temporally Grounding Natural Sentence in Video}. In
  \bibinfo{booktitle}{\emph{Proceedings of the Conference on Empirical Methods
  in Natural Language Processing}}. \bibinfo{publisher}{ACL},
  \bibinfo{pages}{162--171}.
\newblock


\bibitem[\protect\citeauthoryear{Chen, Ma, Chen, Jie, and Luo}{Chen
  et~al\mbox{.}}{2019a}]%
        {chen2019localizing}
\bibfield{author}{\bibinfo{person}{Jingyuan Chen}, \bibinfo{person}{Lin Ma},
  \bibinfo{person}{Xinpeng Chen}, \bibinfo{person}{Zequn Jie}, {and}
  \bibinfo{person}{Jiebo Luo}.} \bibinfo{year}{2019}\natexlab{a}.
\newblock \showarticletitle{Localizing Natural Language in Videos}. In
  \bibinfo{booktitle}{\emph{Proceedings of the American Association for
  Artificial Intelligence}}.
\newblock


\bibitem[\protect\citeauthoryear{Chen, Lu, Tang, Xiao, Zhang, Tan, and Li}{Chen
  et~al\mbox{.}}{2020a}]%
        {chenrethinking}
\bibfield{author}{\bibinfo{person}{Long Chen}, \bibinfo{person}{Chujie Lu},
  \bibinfo{person}{Siliang Tang}, \bibinfo{person}{Jun Xiao},
  \bibinfo{person}{Dong Zhang}, \bibinfo{person}{Chilie Tan}, {and}
  \bibinfo{person}{Xiaolin Li}.} \bibinfo{year}{2020}\natexlab{a}.
\newblock \showarticletitle{Rethinking the Bottom-Up Framework for Query-based
  Video Localization}. In \bibinfo{booktitle}{\emph{Proceedings of the American
  Association for Artificial Intelligence}}.
\newblock


\bibitem[\protect\citeauthoryear{Chen, Ma, Luo, Tang, and Wong}{Chen
  et~al\mbox{.}}{2020b}]%
        {chen2020look}
\bibfield{author}{\bibinfo{person}{Zhenfang Chen}, \bibinfo{person}{Lin Ma},
  \bibinfo{person}{Wenhan Luo}, \bibinfo{person}{Peng Tang}, {and}
  \bibinfo{person}{Kwan-Yee~K Wong}.} \bibinfo{year}{2020}\natexlab{b}.
\newblock \showarticletitle{Look Closer to Ground Better: Weakly-Supervised
  Temporal Grounding of Sentence in Video}.
\newblock \bibinfo{journal}{\emph{arXiv preprint arXiv:2001.09308}}
  (\bibinfo{year}{2020}).
\newblock


\bibitem[\protect\citeauthoryear{Chen, Ma, Luo, and Wong}{Chen
  et~al\mbox{.}}{2019b}]%
        {chen2019weakly}
\bibfield{author}{\bibinfo{person}{Zhenfang Chen}, \bibinfo{person}{Lin Ma},
  \bibinfo{person}{Wenhan Luo}, {and} \bibinfo{person}{Kwan-Yee~K Wong}.}
  \bibinfo{year}{2019}\natexlab{b}.
\newblock \showarticletitle{Weakly-Supervised Spatio-Temporally Grounding
  Natural Sentence in Video}. In \bibinfo{booktitle}{\emph{Proceedings of the
  Conference on the Association for Computational Linguistics}}.
\newblock


\bibitem[\protect\citeauthoryear{Chung, Gulcehre, Cho, and Bengio}{Chung
  et~al\mbox{.}}{2014}]%
        {chung2014empirical}
\bibfield{author}{\bibinfo{person}{Junyoung Chung}, \bibinfo{person}{Caglar
  Gulcehre}, \bibinfo{person}{KyungHyun Cho}, {and} \bibinfo{person}{Yoshua
  Bengio}.} \bibinfo{year}{2014}\natexlab{}.
\newblock \showarticletitle{Empirical evaluation of gated recurrent neural
  networks on sequence modeling}. In \bibinfo{booktitle}{\emph{Advances in
  Neural Information Processing Systems}}.
\newblock


\bibitem[\protect\citeauthoryear{Duan, Huang, Gan, Wang, Zhu, and Huang}{Duan
  et~al\mbox{.}}{2018}]%
        {duan2018weakly}
\bibfield{author}{\bibinfo{person}{Xuguang Duan}, \bibinfo{person}{Wenbing
  Huang}, \bibinfo{person}{Chuang Gan}, \bibinfo{person}{Jingdong Wang},
  \bibinfo{person}{Wenwu Zhu}, {and} \bibinfo{person}{Junzhou Huang}.}
  \bibinfo{year}{2018}\natexlab{}.
\newblock \showarticletitle{Weakly supervised dense event captioning in
  videos}. In \bibinfo{booktitle}{\emph{Advances in Neural Information
  Processing Systems}}. \bibinfo{pages}{3059--3069}.
\newblock


\bibitem[\protect\citeauthoryear{Duchi, Hazan, and Singer}{Duchi
  et~al\mbox{.}}{2011}]%
        {duchi2011adaptive}
\bibfield{author}{\bibinfo{person}{John Duchi}, \bibinfo{person}{Elad Hazan},
  {and} \bibinfo{person}{Yoram Singer}.} \bibinfo{year}{2011}\natexlab{}.
\newblock \showarticletitle{Adaptive subgradient methods for online learning
  and stochastic optimization}.
\newblock \bibinfo{journal}{\emph{Journal of Machine Learning Research}}
  \bibinfo{volume}{12}, \bibinfo{number}{Jul} (\bibinfo{year}{2011}),
  \bibinfo{pages}{2121--2159}.
\newblock


\bibitem[\protect\citeauthoryear{Gao, Sun, Yang, and Nevatia}{Gao
  et~al\mbox{.}}{2017}]%
        {gao2017tall}
\bibfield{author}{\bibinfo{person}{Jiyang Gao}, \bibinfo{person}{Chen Sun},
  \bibinfo{person}{Zhenheng Yang}, {and} \bibinfo{person}{Ram Nevatia}.}
  \bibinfo{year}{2017}\natexlab{}.
\newblock \showarticletitle{{TALL:} Temporal Activity Localization via Language
  Query}. In \bibinfo{booktitle}{\emph{Proceedings of the IEEE International
  Conference on Computer Vision}}. IEEE, \bibinfo{pages}{5277--5285}.
\newblock


\bibitem[\protect\citeauthoryear{Gao, Davis, Socher, and Xiong}{Gao
  et~al\mbox{.}}{2019}]%
        {gao2019wslln}
\bibfield{author}{\bibinfo{person}{Mingfei Gao}, \bibinfo{person}{Larry~S
  Davis}, \bibinfo{person}{Richard Socher}, {and} \bibinfo{person}{Caiming
  Xiong}.} \bibinfo{year}{2019}\natexlab{}.
\newblock \showarticletitle{WSLLN: Weakly Supervised Natural Language
  Localization Networks}.
\newblock \bibinfo{journal}{\emph{Proceedings of the Conference on Empirical
  Methods in Natural Language Processing}} (\bibinfo{year}{2019}).
\newblock


\bibitem[\protect\citeauthoryear{He, Zhao, Huang, Li, Liu, and Wen}{He
  et~al\mbox{.}}{2019}]%
        {he2019read}
\bibfield{author}{\bibinfo{person}{Dongliang He}, \bibinfo{person}{Xiang Zhao},
  \bibinfo{person}{Jizhou Huang}, \bibinfo{person}{Fu Li},
  \bibinfo{person}{Xiao Liu}, {and} \bibinfo{person}{Shilei Wen}.}
  \bibinfo{year}{2019}\natexlab{}.
\newblock \showarticletitle{Read, watch, and move: Reinforcement learning for
  temporally grounding natural language descriptions in videos}. In
  \bibinfo{booktitle}{\emph{Proceedings of the American Association for
  Artificial Intelligence}}, Vol.~\bibinfo{volume}{33}.
  \bibinfo{pages}{8393--8400}.
\newblock


\bibitem[\protect\citeauthoryear{Hendricks, Wang, Shechtman, Sivic, Darrell,
  and Russell}{Hendricks et~al\mbox{.}}{2017}]%
        {hendricks2017localizing}
\bibfield{author}{\bibinfo{person}{Lisa~Anne Hendricks},
  \bibinfo{person}{Oliver Wang}, \bibinfo{person}{Eli Shechtman},
  \bibinfo{person}{Josef Sivic}, \bibinfo{person}{Trevor Darrell}, {and}
  \bibinfo{person}{Bryan Russell}.} \bibinfo{year}{2017}\natexlab{}.
\newblock \showarticletitle{Localizing moments in video with natural language}.
  In \bibinfo{booktitle}{\emph{Proceedings of the IEEE International Conference
  on Computer Vision}}. \bibinfo{pages}{5803--5812}.
\newblock


\bibitem[\protect\citeauthoryear{Hendricks, Wang, Shechtman, Sivic, Darrell,
  and Russell}{Hendricks et~al\mbox{.}}{2018}]%
        {hendricks2018localizing}
\bibfield{author}{\bibinfo{person}{Lisa~Anne Hendricks},
  \bibinfo{person}{Oliver Wang}, \bibinfo{person}{Eli Shechtman},
  \bibinfo{person}{Josef Sivic}, \bibinfo{person}{Trevor Darrell}, {and}
  \bibinfo{person}{Bryan Russell}.} \bibinfo{year}{2018}\natexlab{}.
\newblock \showarticletitle{Localizing Moments in Video with Temporal
  Language}. In \bibinfo{booktitle}{\emph{Proceedings of the Conference on
  Empirical Methods in Natural Language Processing}}. \bibinfo{publisher}{ACL},
  \bibinfo{pages}{1380--1390}.
\newblock


\bibitem[\protect\citeauthoryear{Karpathy and Fei-Fei}{Karpathy and
  Fei-Fei}{2015}]%
        {karpathy2015deep}
\bibfield{author}{\bibinfo{person}{Andrej Karpathy} {and} \bibinfo{person}{Li
  Fei-Fei}.} \bibinfo{year}{2015}\natexlab{}.
\newblock \showarticletitle{Deep visual-semantic alignments for generating
  image descriptions}. In \bibinfo{booktitle}{\emph{Proceedings of the IEEE
  Conference on Computer Vision and Pattern Recognition}}.
  \bibinfo{pages}{3128--3137}.
\newblock


\bibitem[\protect\citeauthoryear{Lin, Zhao, Zhang, Wang, and Liu}{Lin
  et~al\mbox{.}}{2020a}]%
        {lin2019weakly}
\bibfield{author}{\bibinfo{person}{Zhijie Lin}, \bibinfo{person}{Zhou Zhao},
  \bibinfo{person}{Zhu Zhang}, \bibinfo{person}{Qi Wang}, {and}
  \bibinfo{person}{Huasheng Liu}.} \bibinfo{year}{2020}\natexlab{a}.
\newblock \showarticletitle{Weakly-Supervised Video Moment Retrieval via
  Semantic Completion Network}. In \bibinfo{booktitle}{\emph{Proceedings of the
  American Association for Artificial Intelligence}}.
\newblock


\bibitem[\protect\citeauthoryear{Lin, Zhao, Zhang, Zhang, and Cai}{Lin
  et~al\mbox{.}}{2020b}]%
        {lin2020moment}
\bibfield{author}{\bibinfo{person}{Zhijie Lin}, \bibinfo{person}{Zhou Zhao},
  \bibinfo{person}{Zhu Zhang}, \bibinfo{person}{Zijian Zhang}, {and}
  \bibinfo{person}{Deng Cai}.} \bibinfo{year}{2020}\natexlab{b}.
\newblock \showarticletitle{Moment Retrieval via Cross-Modal Interaction
  Networks With Query Reconstruction}.
\newblock \bibinfo{journal}{\emph{IEEE Transactions on Image Processing}}
  \bibinfo{volume}{29} (\bibinfo{year}{2020}), \bibinfo{pages}{3750--3762}.
\newblock


\bibitem[\protect\citeauthoryear{Liu, Jiang, and Wang}{Liu
  et~al\mbox{.}}{2019}]%
        {liu2019completeness}
\bibfield{author}{\bibinfo{person}{Daochang Liu}, \bibinfo{person}{Tingting
  Jiang}, {and} \bibinfo{person}{Yizhou Wang}.}
  \bibinfo{year}{2019}\natexlab{}.
\newblock \showarticletitle{Completeness modeling and context separation for
  weakly supervised temporal action localization}. In
  \bibinfo{booktitle}{\emph{Proceedings of the IEEE Conference on Computer
  Vision and Pattern Recognition}}. \bibinfo{pages}{1298--1307}.
\newblock


\bibitem[\protect\citeauthoryear{Liu, Wang, Nie, He, Chen, and Chua}{Liu
  et~al\mbox{.}}{2018a}]%
        {liu2018attentive}
\bibfield{author}{\bibinfo{person}{Meng Liu}, \bibinfo{person}{Xiang Wang},
  \bibinfo{person}{Liqiang Nie}, \bibinfo{person}{Xiangnan He},
  \bibinfo{person}{Baoquan Chen}, {and} \bibinfo{person}{Tat-Seng Chua}.}
  \bibinfo{year}{2018}\natexlab{a}.
\newblock \showarticletitle{Attentive moment retrieval in videos}. In
  \bibinfo{booktitle}{\emph{Proceedings of the International ACM SIGIR
  Conference on Research and Development in Information Retrieval}}. ACM,
  \bibinfo{pages}{15--24}.
\newblock


\bibitem[\protect\citeauthoryear{Liu, Wang, Nie, Tian, Chen, and Chua}{Liu
  et~al\mbox{.}}{2018b}]%
        {liu2018cross}
\bibfield{author}{\bibinfo{person}{Meng Liu}, \bibinfo{person}{Xiang Wang},
  \bibinfo{person}{Liqiang Nie}, \bibinfo{person}{Qi Tian},
  \bibinfo{person}{Baoquan Chen}, {and} \bibinfo{person}{Tat-Seng Chua}.}
  \bibinfo{year}{2018}\natexlab{b}.
\newblock \showarticletitle{Cross-modal Moment Localization in Videos}. In
  \bibinfo{booktitle}{\emph{Proceedings of the ACM International Conference on
  Multimedia}}. \bibinfo{publisher}{ACM}, \bibinfo{pages}{843--851}.
\newblock


\bibitem[\protect\citeauthoryear{Mithun, Paul, and Roy-Chowdhury}{Mithun
  et~al\mbox{.}}{2019}]%
        {mithun2019weakly}
\bibfield{author}{\bibinfo{person}{Niluthpol~Chowdhury Mithun},
  \bibinfo{person}{Sujoy Paul}, {and} \bibinfo{person}{Amit~K Roy-Chowdhury}.}
  \bibinfo{year}{2019}\natexlab{}.
\newblock \showarticletitle{Weakly supervised video moment retrieval from text
  queries}. In \bibinfo{booktitle}{\emph{Proceedings of the IEEE Conference on
  Computer Vision and Pattern Recognition}}. \bibinfo{pages}{11592--11601}.
\newblock


\bibitem[\protect\citeauthoryear{Nguyen, Liu, Prasad, and Han}{Nguyen
  et~al\mbox{.}}{2018}]%
        {nguyen2018weakly}
\bibfield{author}{\bibinfo{person}{Phuc Nguyen}, \bibinfo{person}{Ting Liu},
  \bibinfo{person}{Gautam Prasad}, {and} \bibinfo{person}{Bohyung Han}.}
  \bibinfo{year}{2018}\natexlab{}.
\newblock \showarticletitle{Weakly supervised action localization by sparse
  temporal pooling network}. In \bibinfo{booktitle}{\emph{Proceedings of the
  IEEE Conference on Computer Vision and Pattern Recognition}}.
  \bibinfo{pages}{6752--6761}.
\newblock


\bibitem[\protect\citeauthoryear{Pennington, Socher, and Manning}{Pennington
  et~al\mbox{.}}{2014}]%
        {pennington2014glove}
\bibfield{author}{\bibinfo{person}{Jeffrey Pennington},
  \bibinfo{person}{Richard Socher}, {and} \bibinfo{person}{Christopher
  Manning}.} \bibinfo{year}{2014}\natexlab{}.
\newblock \showarticletitle{Glove: Global vectors for word representation}. In
  \bibinfo{booktitle}{\emph{Proceedings of the Conference on Empirical Methods
  in Natural Language Processing}}. \bibinfo{pages}{1532--1543}.
\newblock


\bibitem[\protect\citeauthoryear{Ren, He, Girshick, and Sun}{Ren
  et~al\mbox{.}}{2015}]%
        {ren2015faster}
\bibfield{author}{\bibinfo{person}{Shaoqing Ren}, \bibinfo{person}{Kaiming He},
  \bibinfo{person}{Ross Girshick}, {and} \bibinfo{person}{Jian Sun}.}
  \bibinfo{year}{2015}\natexlab{}.
\newblock \showarticletitle{Faster r-cnn: Towards real-time object detection
  with region proposal networks}. In \bibinfo{booktitle}{\emph{Advances in
  Neural Information Processing Systems}}. \bibinfo{pages}{91--99}.
\newblock


\bibitem[\protect\citeauthoryear{Shou, Chan, Zareian, Miyazawa, and Chang}{Shou
  et~al\mbox{.}}{2017}]%
        {shou2017cdc}
\bibfield{author}{\bibinfo{person}{Zheng Shou}, \bibinfo{person}{Jonathan
  Chan}, \bibinfo{person}{Alireza Zareian}, \bibinfo{person}{Kazuyuki
  Miyazawa}, {and} \bibinfo{person}{Shih-Fu Chang}.}
  \bibinfo{year}{2017}\natexlab{}.
\newblock \showarticletitle{Cdc: Convolutional-de-convolutional networks for
  precise temporal action localization in untrimmed videos}. In
  \bibinfo{booktitle}{\emph{Proceedings of the IEEE Conference on Computer
  Vision and Pattern Recognition}}. IEEE, \bibinfo{pages}{1417--1426}.
\newblock


\bibitem[\protect\citeauthoryear{Shou, Gao, Zhang, Miyazawa, and Chang}{Shou
  et~al\mbox{.}}{2018}]%
        {shou2018autoloc}
\bibfield{author}{\bibinfo{person}{Zheng Shou}, \bibinfo{person}{Hang Gao},
  \bibinfo{person}{Lei Zhang}, \bibinfo{person}{Kazuyuki Miyazawa}, {and}
  \bibinfo{person}{Shih-Fu Chang}.} \bibinfo{year}{2018}\natexlab{}.
\newblock \showarticletitle{Autoloc: Weakly-supervised temporal action
  localization in untrimmed videos}. In \bibinfo{booktitle}{\emph{Proceedings
  of the European Conference on Computer Vision}}. \bibinfo{pages}{154--171}.
\newblock


\bibitem[\protect\citeauthoryear{Shou, Wang, and Chang}{Shou
  et~al\mbox{.}}{2016}]%
        {shou2016temporal}
\bibfield{author}{\bibinfo{person}{Zheng Shou}, \bibinfo{person}{Dongang Wang},
  {and} \bibinfo{person}{Shih-Fu Chang}.} \bibinfo{year}{2016}\natexlab{}.
\newblock \showarticletitle{Temporal action localization in untrimmed videos
  via multi-stage cnns}. In \bibinfo{booktitle}{\emph{Proceedings of the IEEE
  Conference on Computer Vision and Pattern Recognition}}.
  \bibinfo{pages}{1049--1058}.
\newblock


\bibitem[\protect\citeauthoryear{Sigurdsson, Varol, Wang, Farhadi, Laptev, and
  Gupta}{Sigurdsson et~al\mbox{.}}{2016}]%
        {sigurdssonhollywood}
\bibfield{author}{\bibinfo{person}{Gunnar~A Sigurdsson},
  \bibinfo{person}{G{\"u}l Varol}, \bibinfo{person}{Xiaolong Wang},
  \bibinfo{person}{Ali Farhadi}, \bibinfo{person}{Ivan Laptev}, {and}
  \bibinfo{person}{Abhinav Gupta}.} \bibinfo{year}{2016}\natexlab{}.
\newblock \showarticletitle{Hollywood in Homes: Crowdsourcing Data Collection
  for Activity Understanding}. In \bibinfo{booktitle}{\emph{Proceedings of the
  European Conference on Computer Vision}}.
\newblock


\bibitem[\protect\citeauthoryear{Tran, Bourdev, Fergus, Torresani, and
  Paluri}{Tran et~al\mbox{.}}{2015}]%
        {tran2015learning}
\bibfield{author}{\bibinfo{person}{Du Tran}, \bibinfo{person}{Lubomir Bourdev},
  \bibinfo{person}{Rob Fergus}, \bibinfo{person}{Lorenzo Torresani}, {and}
  \bibinfo{person}{Manohar Paluri}.} \bibinfo{year}{2015}\natexlab{}.
\newblock \showarticletitle{Learning spatiotemporal features with 3d
  convolutional networks}. In \bibinfo{booktitle}{\emph{Proceedings of the IEEE
  International Conference on Computer Vision}}. \bibinfo{pages}{4489--4497}.
\newblock


\bibitem[\protect\citeauthoryear{Wang, Xiong, Lin, and Van~Gool}{Wang
  et~al\mbox{.}}{2017}]%
        {wang2017untrimmednets}
\bibfield{author}{\bibinfo{person}{Limin Wang}, \bibinfo{person}{Yuanjun
  Xiong}, \bibinfo{person}{Dahua Lin}, {and} \bibinfo{person}{Luc Van~Gool}.}
  \bibinfo{year}{2017}\natexlab{}.
\newblock \showarticletitle{Untrimmednets for weakly supervised action
  recognition and detection}. In \bibinfo{booktitle}{\emph{Proceedings of the
  IEEE Conference on Computer Vision and Pattern Recognition}}.
\newblock


\bibitem[\protect\citeauthoryear{Wang, Huang, and Wang}{Wang
  et~al\mbox{.}}{2019}]%
        {wang2019language}
\bibfield{author}{\bibinfo{person}{Weining Wang}, \bibinfo{person}{Yan Huang},
  {and} \bibinfo{person}{Liang Wang}.} \bibinfo{year}{2019}\natexlab{}.
\newblock \showarticletitle{Language-Driven Temporal Activity Localization: A
  Semantic Matching Reinforcement Learning Model}. In
  \bibinfo{booktitle}{\emph{Proceedings of the IEEE Conference on Computer
  Vision and Pattern Recognition}}. \bibinfo{pages}{334--343}.
\newblock


\bibitem[\protect\citeauthoryear{Xu, He, Sigal, Sclaroff, and Saenko}{Xu
  et~al\mbox{.}}{2019}]%
        {xu2019multilevel}
\bibfield{author}{\bibinfo{person}{Huijuan Xu}, \bibinfo{person}{Kun He},
  \bibinfo{person}{L Sigal}, \bibinfo{person}{S Sclaroff}, {and}
  \bibinfo{person}{K Saenko}.} \bibinfo{year}{2019}\natexlab{}.
\newblock \showarticletitle{Multilevel Language and Vision Integration for
  Text-to-Clip Retrieval}. In \bibinfo{booktitle}{\emph{Proceedings of the
  American Association for Artificial Intelligence}}, Vol.~\bibinfo{volume}{2}.
  \bibinfo{pages}{7}.
\newblock


\bibitem[\protect\citeauthoryear{Yu, Ren, Li, Yan, Xu, and Yuan}{Yu
  et~al\mbox{.}}{2019}]%
        {yu2019temporal}
\bibfield{author}{\bibinfo{person}{Tan Yu}, \bibinfo{person}{Zhou Ren},
  \bibinfo{person}{Yuncheng Li}, \bibinfo{person}{Enxu Yan},
  \bibinfo{person}{Ning Xu}, {and} \bibinfo{person}{Junsong Yuan}.}
  \bibinfo{year}{2019}\natexlab{}.
\newblock \showarticletitle{Temporal structure mining for weakly supervised
  action detection}. In \bibinfo{booktitle}{\emph{Proceedings of the IEEE
  International Conference on Computer Vision}}. \bibinfo{pages}{5522--5531}.
\newblock


\bibitem[\protect\citeauthoryear{Yuan, Ma, Wang, Liu, and Zhu}{Yuan
  et~al\mbox{.}}{2019}]%
        {yuan2019semantic}
\bibfield{author}{\bibinfo{person}{Yitian Yuan}, \bibinfo{person}{Lin Ma},
  \bibinfo{person}{Jingwen Wang}, \bibinfo{person}{Wei Liu}, {and}
  \bibinfo{person}{Wenwu Zhu}.} \bibinfo{year}{2019}\natexlab{}.
\newblock \showarticletitle{Semantic Conditioned Dynamic Modulation for
  Temporal Sentence Grounding in Videos}. In \bibinfo{booktitle}{\emph{Advances
  in Neural Information Processing Systems}}. \bibinfo{pages}{534--544}.
\newblock


\bibitem[\protect\citeauthoryear{Zeng, Huang, Tan, Rong, Zhao, Huang, and
  Gan}{Zeng et~al\mbox{.}}{2019}]%
        {zeng2019graph}
\bibfield{author}{\bibinfo{person}{Runhao Zeng}, \bibinfo{person}{Wenbing
  Huang}, \bibinfo{person}{Mingkui Tan}, \bibinfo{person}{Yu Rong},
  \bibinfo{person}{Peilin Zhao}, \bibinfo{person}{Junzhou Huang}, {and}
  \bibinfo{person}{Chuang Gan}.} \bibinfo{year}{2019}\natexlab{}.
\newblock \showarticletitle{Graph Convolutional Networks for Temporal Action
  Localization}. In \bibinfo{booktitle}{\emph{Proceedings of the IEEE
  International Conference on Computer Vision}}.
\newblock


\bibitem[\protect\citeauthoryear{Zhang, Dai, Wang, Wang, and Davis}{Zhang
  et~al\mbox{.}}{2019a}]%
        {zhang2019man}
\bibfield{author}{\bibinfo{person}{Da Zhang}, \bibinfo{person}{Xiyang Dai},
  \bibinfo{person}{Xin Wang}, \bibinfo{person}{Yuan-Fang Wang}, {and}
  \bibinfo{person}{Larry~S Davis}.} \bibinfo{year}{2019}\natexlab{a}.
\newblock \showarticletitle{Man: Moment alignment network for natural language
  moment retrieval via iterative graph adjustment}. In
  \bibinfo{booktitle}{\emph{Proceedings of the IEEE Conference on Computer
  Vision and Pattern Recognition}}. \bibinfo{pages}{1247--1257}.
\newblock


\bibitem[\protect\citeauthoryear{Zhang, Peng, Fu, and Luo}{Zhang
  et~al\mbox{.}}{2020a}]%
        {zhang2019learning}
\bibfield{author}{\bibinfo{person}{Songyang Zhang}, \bibinfo{person}{Houwen
  Peng}, \bibinfo{person}{Jianlong Fu}, {and} \bibinfo{person}{Jiebo Luo}.}
  \bibinfo{year}{2020}\natexlab{a}.
\newblock \showarticletitle{Learning 2D Temporal Adjacent Networks for Moment
  Localization with Natural Language}. In \bibinfo{booktitle}{\emph{Proceedings
  of the American Association for Artificial Intelligence}}.
\newblock


\bibitem[\protect\citeauthoryear{Zhang, Lin, Zhao, and Xiao}{Zhang
  et~al\mbox{.}}{2019b}]%
        {zhang2019cross}
\bibfield{author}{\bibinfo{person}{Zhu Zhang}, \bibinfo{person}{Zhijie Lin},
  \bibinfo{person}{Zhou Zhao}, {and} \bibinfo{person}{Zhenxin Xiao}.}
  \bibinfo{year}{2019}\natexlab{b}.
\newblock \showarticletitle{Cross-modal interaction networks for query-based
  moment retrieval in videos}. In \bibinfo{booktitle}{\emph{Proceedings of the
  International ACM SIGIR Conference on Research and Development in Information
  Retrieval}}. \bibinfo{pages}{655--664}.
\newblock


\bibitem[\protect\citeauthoryear{Zhang, Zhao, Lin, Huai, and Yuan}{Zhang
  et~al\mbox{.}}{2020b}]%
        {zhang2020object}
\bibfield{author}{\bibinfo{person}{Zhu Zhang}, \bibinfo{person}{Zhou Zhao},
  \bibinfo{person}{Zhijie Lin}, \bibinfo{person}{Baoxing Huai}, {and}
  \bibinfo{person}{Jing Yuan}.} \bibinfo{year}{2020}\natexlab{b}.
\newblock \showarticletitle{Object-Aware Multi-Branch Relation Networks for
  Spatio-Temporal Video Grounding}. In \bibinfo{booktitle}{\emph{Proceedings of
  the International Joint Conference on Artificial Intelligence}}.
  \bibinfo{pages}{1069--1075}.
\newblock


\bibitem[\protect\citeauthoryear{Zhang, Zhao, Lin, Song, and Cai}{Zhang
  et~al\mbox{.}}{2019c}]%
        {zhang2019localizing}
\bibfield{author}{\bibinfo{person}{Zhu Zhang}, \bibinfo{person}{Zhou Zhao},
  \bibinfo{person}{Zhijie Lin}, \bibinfo{person}{Jingkuan Song}, {and}
  \bibinfo{person}{Deng Cai}.} \bibinfo{year}{2019}\natexlab{c}.
\newblock \showarticletitle{Localizing Unseen Activities in Video via Image
  Query}. In \bibinfo{booktitle}{\emph{Proceedings of the International Joint
  Conference on Artificial Intelligence}}.
\newblock


\bibitem[\protect\citeauthoryear{Zhang, Zhao, Zhao, Wang, Liu, and Gao}{Zhang
  et~al\mbox{.}}{2020c}]%
        {zhang2020does}
\bibfield{author}{\bibinfo{person}{Zhu Zhang}, \bibinfo{person}{Zhou Zhao},
  \bibinfo{person}{Yang Zhao}, \bibinfo{person}{Qi Wang},
  \bibinfo{person}{Huasheng Liu}, {and} \bibinfo{person}{Lianli Gao}.}
  \bibinfo{year}{2020}\natexlab{c}.
\newblock \showarticletitle{Where Does It Exist: Spatio-Temporal Video
  Grounding for Multi-Form Sentences}. In \bibinfo{booktitle}{\emph{Proceedings
  of the IEEE Conference on Computer Vision and Pattern Recognition}}.
  \bibinfo{pages}{10668--10677}.
\newblock


\bibitem[\protect\citeauthoryear{Zhao, Xiong, Wang, Wu, Tang, and Lin}{Zhao
  et~al\mbox{.}}{2017}]%
        {zhao2017temporal}
\bibfield{author}{\bibinfo{person}{Yue Zhao}, \bibinfo{person}{Yuanjun Xiong},
  \bibinfo{person}{Limin Wang}, \bibinfo{person}{Zhirong Wu},
  \bibinfo{person}{Xiaoou Tang}, {and} \bibinfo{person}{Dahua Lin}.}
  \bibinfo{year}{2017}\natexlab{}.
\newblock \showarticletitle{Temporal action detection with structured segment
  networks}. In \bibinfo{booktitle}{\emph{Proceedings of the IEEE International
  Conference on Computer Vision}}.
\newblock


\end{thebibliography}

%%
%% If your work has an appendix, this is the place to put it.
% \appendix

% \section{Research Methods}

% \subsection{Part One}

% Lorem ipsum dolor sit amet, consectetur adipiscing elit. Morbi
% malesuada, quam in pulvinar varius, metus nunc fermentum urna, id
% sollicitudin purus odio sit amet enim. Aliquam ullamcorper eu ipsum
% vel mollis. Curabitur quis dictum nisl. Phasellus vel semper risus, et
% lacinia dolor. Integer ultricies commodo sem nec semper.

% \subsection{Part Two}

% Etiam commodo feugiat nisl pulvinar pellentesque. Etiam auctor sodales
% ligula, non varius nibh pulvinar semper. Suspendisse nec lectus non
% ipsum convallis congue hendrerit vitae sapien. Donec at laoreet
% eros. Vivamus non purus placerat, scelerisque diam eu, cursus
% ante. Etiam aliquam tortor auctor efficitur mattis.

% \section{Online Resources}

% Nam id fermentum dui. Suspendisse sagittis tortor a nulla mollis, in
% pulvinar ex pretium. Sed interdum orci quis metus euismod, et sagittis
% enim maximus. Vestibulum gravida massa ut felis suscipit
% congue. Quisque mattis elit a risus ultrices commodo venenatis eget
% dui. Etiam sagittis eleifend elementum.

% Nam interdum magna at lectus dignissim, ac dignissim lorem
% rhoncus. Maecenas eu arcu ac neque placerat aliquam. Nunc pulvinar
% massa et mattis lacinia.

\end{document}